%% file: header.tex
\documentclass[conference]{IEEEtran}

\usepackage{subfig}
\usepackage{graphicx}
\usepackage{wrapfig}
\usepackage{lipsum}  
\usepackage{fancyvrb}
\usepackage{url}
\usepackage{amsmath}
\usepackage{wrapfig}
\usepackage{enumerate}
\usepackage{amsmath,amssymb,amsthm}
\usepackage[linesnumbered,algoruled,boxed,lined]{algorithm2e}
\usepackage{xcolor}

\newcommand{\revision}[1]{}

\usepackage{dashbox}

\definecolor{myblue}{rgb}{0,0,128}

\SetCommentSty{mycommfont}

\newcommand{\myparam}[1]{\left( {#1} \right)}

\newcommand{\myrect}[1]{\left\lbrack #1 \right\rbrack}

\newcommand\textproc{\textsc}

\newcommand{\dotavg}[1]{\frac{{\bf 1}_n}{#1}}

\newcommand{\norm}[1]{{\left\lVert {#1}\right\rVert}^2}

\newcommand{\normF}[1]{\left\lVert #1 \right\rVert_F^2}

\newcommand{\E}{\rm I\kern-.3em E}
\newcommand{\Var}[1]{\mathrm{Var}\left\lbrack #1 \right\rbrack}

\SetKwInOut{Require}{Require}

\begin{document}

\title{Adaptive Periodic Averaging: A Practical Approach to Reducing Communication in Distributed Learning}

\author{
\IEEEauthorblockN{Peng Jiang}
\IEEEauthorblockA{
\textit{The University of Iowa}\\
peng-jiang@uiowa.edu}
\and 
\IEEEauthorblockN{Gagan Agrawal}
\IEEEauthorblockA{
\textit{Augusta University}\\
gagrawal@augusta.edu}
}

\maketitle
\thispagestyle{plain}
\pagestyle{plain}

\input{text/abstract}

\input{text/introduction}

\input{text/Background}

\input{text/analysis}

\input{text/algorithm}

\input{text/experiments} 
\input{text/discuss} 
\input{text/related}

\input{text/conclusion}

\bibliographystyle{IEEEtran}
\bibliography{BIB/jp}


\end{document}

%% file: text/abstract.tex
\begin{abstract}
Stochastic Gradient Descent (SGD) is the key learning algorithm for many  machine learning  tasks. 
Because of its computational costs,  there is a growing interest in accelerating SGD on  HPC 
resources like GPU clusters. 
However,  the performance of  parallel SGD is still bottlenecked by the high communication costs 
 even with a fast connection among the machines. 
A simple  approach to alleviating this problem, used in many existing efforts, 
is  to  perform communication every few iterations,   using 
a constant  {\em averaging period}. 
In this paper, we show that the optimal averaging period in terms of convergence  and  communication cost  
is not a constant, but instead varies over the course of the execution. 
Specifically, we observe that reducing  the variance of model parameters among the computing nodes is 
critical to  
the convergence of  periodic parameter averaging SGD.  
 Given a fixed  communication budget,  we show that it is more beneficial to synchronize more frequently in early iterations to reduce the initial large variance and synchronize less frequently in the 
 later phase of the training process.
We propose a practical algorithm, named  ADaptive Periodic parameter averaging SGD (ADPSGD),  to achieve a smaller overall  
variance of model  parameters,  and thus better convergence compared
with the Constant Periodic parameter averaging SGD (CPSGD). 
We  evaluate our method with several image classification benchmarks, and 
show that our ADPSGD indeed achieves smaller training losses and higher test accuracies with smaller communication cost 
compared with CPSGD. 
Compared with gradient-quantization SGD, we show that our algorithm achieves faster convergence with only half of the communication. 
Compared with full-communication SGD, our ADPSGD achieves $1.14$x $\sim$ $1.27$x  speedups with a  
100Gbps connection among computing nodes, and the speedups increase to $1.46$x $\sim$ $1.95$x with  a
10Gbps connection. 
\end{abstract}


%% file: text/introduction.tex
\section{Introduction}
\label{sec:intro}

As machine learning today is involving  training deeper and wider neural networks with larger   
datasets, the compute and memory requirements  for  these {\em deep learning} tasks have been increasing. 
This has led to great interest in scaling SGD on  parallel 
 systems~\cite{Dean:2012:LSD:2999134.2999271, Dekel:2012:ODO:2503308.2188391}. 
In fact,  parallel  training is supported in almost all of today's mainstream deep learning frameworks such as Tensorflow~\cite{DBLP:journals/corr/AbadiABBCCCDDDG16}, PyTorch~\cite{NIPS2019_9015}, MXNet~\cite{DBLP:journals/corr/ChenLLLWWXXZZ15}, CNTK~\cite{Seide:2016:CMO:2939672.2945397}, and Caffe~\cite{Jia:2014:CCA:2647868.2654889}.    

The most widely adopted approach to distributed training is a {\em data-parallel} SGD. 
The idea is that each machine holds a copy of the entire model and computes stochastic gradients with  
local {\em mini-batches}.  
The  local model parameters or gradients are synchronized frequently to achieve a global consensus of the  
learned model. 
Though the  parallelization is straightforward, a naive implementation cannot achieve good  
performance because the  communication  of the gradients  in each iteration is expensive~\cite{NIPS2011_4247}\cite{Li:2014:SDM:2685048.2685095}\cite{Ho:2013:MED:2999611.2999748}\cite{NIPS2014_5597}\cite{Strom2015ScalableDD}\cite{Li2016ScalingDM}.  
As reported in~\cite{NIPS2017_6768}, $50\%$ to $80\%$ of the total execution time is spent on communication  
when training  deep neural networks on 16 GPUs from AWS EC2. 
As we will show  through our  experiments, even on a HPC cluster with a 100Gbps InfiniBand connection, communication can still take up more than $50\%$ of the total execution time for training certain neural networks. 

The study of communication-efficient SGD reduces to  the exploration of communication strategies that can achieve the best trade-off between convergence and communication for a given configuration. 
A common approach is to synchronize the model parameters among the machines once  every few iterations (namely, {\em periodic parameter averaging}). 
This method has several advantages. 
First, it reduces both the bandwidth cost and latency in communication -- in comparison, compression-based methods that we will discuss 
in next section only save the bandwidth. 
Second, it can be easily combined with bandwidth-optimal Allreduce~\cite{10.1016/j.jpdc.2008.09.002}.
Third,  again unlike compression or quantization methods, it requires no or little extra computation.  
Periodic parameter averaging has been adopted in many previous works for accelerating SGD~\cite{JMLR:v15:hazan14a, Johnson:2013:ASG:2999611.2999647, DBLP:journals/corr/SmithFMTJJ16, 2016arXiv160607365Z, 45187, 2017arXiv170206269W}.  
Recent works~\cite{NIPS2018_7519}\cite{DBLP:journals/corr/abs-1708-01012} also show that periodic parameter averaging can achieve the well-known $O(1/\sqrt{MK})$ convergence rate for distributed SGD on non-convex optimization if a ``proper'' averaging period is used (Here, $M$ is the total mini-batch size in each iteration and $K$ is the number of iterations).   

However, all of the work in this area  assumes that a constant averaging period is to be used throughout the training. As we will show in our experiments, training with a constant averaging period of $8$ can lead to noticeable decrease in convergence and accuracy, which indicates that a naive implementation of periodic parameter averaging SGD is not effective.  
Moreover, Zhou {\em et al.}~\cite{DBLP:journals/corr/abs-1708-01012} have shown  that it is hard to determine the optimal averaging period in practice, and none of the previous works give a specific algorithm to determine this important parameter.


In this paper, we establish both theoretically and empirically that, given a fixed communication budget,  the optimal averaging period for distributed SGD should  be {\em adaptive}. 
We  observe that, when  using a constant averaging period, the variance of model parameters is large  
initially,  but decreases  quickly  over the iterations. 
The large initial variance in constant periodic averaging SGD is  harmful to the  
convergence rate, whereas  fast decrease of the variance  turns out to be  unnecessary for  
achieving the asymptotic $O(1/\sqrt{MK})$ convergence rate. 
Based on this observation, 
we present an algorithm called ADaptive Periodic parameter averaging SGD (ADPSGD) which uses a small averaging period in the beginning and gradually increases it  over the 
iterations.  
When increasing the averaging period, the algorithm keeps the variance of model parameters on different nodes close to a value proportional to the {\em learning  
rate}.
We show that this adaptive periodic averaging strategy maintains the $O(1/\sqrt{MK})$ convergence rate while requiring less communication than constant periodic averaging.  


We evaluate our algorithm on multiple image classification benchmarks.  
Compared with constant periodic parameter averageing SGD, our algorithm achieves smaller training loss and higher  
test accuracy,  while 
requiring smaller communication and total execution time. 
Compared against full-communication SGD,   
our algorithm runs 1.14x to 1.27x faster on 16 Nvidia Tesla P100 GPUs connected by 100Gbps InfiniBand, and 1.46x to 1.95x faster when the connection bandwidth is throttled to 10Gbps.
Compared against single-node SGD, our algorithm achieves linear speedups across 16 nodes due to the saved communication. 
Compared with gradient-quantization SGD, our algorithm achieves faster convergence with only half of the communication.

%% file: text/Background.tex
\section{Background}
This section  provides  background on stochastic gradient descent and constant periodic parameter averaging SGD. 
\subsection{Stochastic Gradient Descent}
Many machine learning problems can be summarized as follows. 
Given that $w$ is a vector of model parameters, the goal is to 
find a value  of $w$  ($w^*$) such that we 
minimize an objective function of the following form:
\begin{equation}
\label{eq:obj1}
	 f(w)\triangleq \frac{1}{N}\sum_{i=1}^{N}{F_i(w)}.
\end{equation}
Here, $N$ is the number of training samples, 
and $F_i(w)$ is the {\em loss function}  on the $i$th sample.  The objective 
function can also be viewed as computing a {\em loss} and 
$w^*$ is considered as the best fit to the training data. 
The loss function can be either {\em convex}  or {\em non-convex}. 
Traditional machine learning models such as linear regression and SVM have convex loss functions  that 
 contain only one global minimum~\cite{ben2001lectures},  whereas  deep neural networks usually have  
non-convex loss functions  that  may contain many local minima~\cite{Goodfellow-et-al-2016}.

\textbf{Algorithm Description.}
{\em Gradient Descent} (GD) is a popular method to find the minimum of a differentiable function. 
Applied to the objective function in (\ref{eq:obj1}), it updates the model parameter $w$ iteratively as:
\begin{equation}
\label{eq:gd}
	 w_{k+1} = w_k - \gamma_k \nabla f(w_k),
\end{equation}
where $\nabla f(w_k) = \frac{1}{N}\sum_{i=1}^{N}{\nabla F_i(w_k)}$ is the gradient of the loss function at point $w_k$, and $\gamma_k$ is the learning rate in the iteration $k$. 
In practice, because the training data can have a large number of samples, it is expensive to compute the accurate gradient at each point. 
Therefore, {\em Stochastic Gradient Descent} (SGD) estimates the  gradient at each point by computing it 
only on one randomly selected sample. 
 Formally,  $\nabla F_i(w_k)$ for a random sample $i$ is used in place of $\nabla f(w_k)$ in (\ref{eq:gd}).  

A tradeoff between GD and pure SGD is {\em mini-batch SGD}, which estimates the gradient at each step with a subset of randomly selected samples. 
More precisely, mini-batch SGD computes the gradient at each point as:
\begin{equation}
\label{eq:eg}
	 \nabla \widetilde{f}(w_k;B_k) = \frac{1}{M}\sum_{i\in B_k}{\nabla F_i(w_k)},
\end{equation} 
where $B_k$ represents a randomly selected mini-batch, and $M$ is the number of samples in the mini-batch.  
The model parameters are updated with the same rule in (\ref{eq:gd}), only with $\nabla f(w_k)$ being replaced by $\nabla \widetilde{f}(w_k;B_k)$.

\textbf{Convergence Rate.}
\label{sec:back2}
 Investigations  of convergence properties of SGD can be traced back more than 60 years  
ago~\cite{robbins1951}. 
Over the years, theories explaining the convergence rates of SGD and its variants on both convex and non-convex optimization have been established~\cite{Robbins_1985, Bottou:1999:OLS:304710.304720, doi:10.1137/070704277, doi:10.1137/120880811}.  


As there are potentially many minima in a non-convex function, the commonly used metric in convergence  
analysis for  such functions is the weighted average of the squared $\ell_2$ norm of all gradients.  Intuitively, a small gradient indicates that the optimization has reached near a minimum.    
It has  been proven that SGD converges at rate $O(1/\sqrt{K})$ for non-convex optimization, which means that  
the average squared gradient norms is smaller than $\epsilon$ after $O(1/\epsilon^2)$ number of iterations~\cite{doi:10.1137/120880811}. 
It has also been shown that mini-batch SGD has convergence rate of $O(1/\sqrt{MK})$ for non-convex  
optimization,  where $M$ is the mini-batch size~\cite{Dekel:2012:ODO:2503308.2188391}. 

The $O(1/\sqrt{MK})$ convergence rate of mini-batch SGD justifies a  data parallel implementation. 
This is  because it indicates that we can use $1/n$ number of iterations to achieve results of the same  
accuracy if we increase the mini-batch size and the learning rate by a factor of $n$. 
However, there is a limit on effective mini-batch size, because the learning rate cannot exceed an  
algorithmic upper bound that  depends on the smoothness of the objective  
function~\cite{Dekel:2012:ODO:2503308.2188391}.  


%% file: text/analysis.tex
\subsection{Constant Periodic Parameter Averaging SGD}
\label{sec:analysis}

Constant periodic parameter averaging has been adopted in many previous works to reduce the communication overhead of distributed SGD~\cite{JMLR:v15:hazan14a, Johnson:2013:ASG:2999611.2999647, DBLP:journals/corr/SmithFMTJJ16, 2016arXiv160607365Z, 45187, 2017arXiv170206269W}. 
Periodic parameter averaging SGD has been shown that it can preserve the asymptotic $O(1/\sqrt{MK})$ convergence rate of full-communication SGD on non-convex optimization~\cite{DBLP:journals/corr/abs-1708-01012}.  
 For the convenience of our discussion in the following sections, we now provide an outline of analysis  of periodic parameter averaging SGD. 
Our goal here is to  establish  the point that the smaller the variance of the model parameters among the computing
nodes, the better the convergence of the algorithm. 
Intuitively, a small variance indicates the trajectories of model parameters on different nodes are not far part and thereby this property leads to better convergence.  
The main idea of adaptive periodic averaging SGD, which we will discuss in \S\ref{sec:adpsgd}, is to minimize the overall variance of model parameters so that we can achieve faster convergence with the same (or even smaller) amount of communication overhead compared with constant periodic averaging SGD.

\begin{algorithm}[t]
\footnotesize
\SetAlgoLined
\Require{the initial model parameters $w_0$}
$w_{0,i}=w_0$\;
\For{$k=0,1,2, \ldots ,K-1$}{
  $\nabla \widetilde{f}(w_{k,i};B_{k,i}) = \frac{1}{m}\sum_{j\in B_{k,i}}{\nabla F_j(w_{k,i})}$\;
  \tcc{Updating parameter variables locally}
  $w_{k+1,i}=w_{k,i}-\gamma_k \nabla \widetilde{f}(w_{k,i};B_{k,i})$\;
  \If{$k \mod p==0$}{
  \tcc{Averaging parameters among nodes}
   $w_{k+1,i}=\frac{1}{n} \sum_{j=1}^{n}{w_{k+1,j}}$\;
  }
 }
 \caption{ (CPSGD) Procedure of constant periodic parameter averaging SGD on the $i$th node}
 \label{alg1}
\end{algorithm} 


\textbf{Algorithm Description.}
We first formulate the procedure of constant periodic parameter averaging SGD (CPSGD) as  Algorithm~\ref{alg1}. 
The algorithm is executed on $n$ nodes for $K$ iterations. 
In each iteration, each node first computes a local stochastic gradient $\nabla \widetilde{f}(w_{k,i};B_{k,i})$ based on a randomly selected mini-batch $B_{k,i}$ (line 3). 
Then, the local model parameters are updated with the stochastic gradient (line 4). 
After the local update, each node checks if $p$ divides the iterate number (line 5). 
If so, the model parameters are synchronized and averaged among all the  nodes (line 6).  

If  $W_k=[w_{k,1}, w_{k,2}, \ldots, w_{k,n}]$ are  the model parameters on all nodes at the beginning of the 
iteration $k$, the main computation in  Algorithm~\ref{alg1} can be expressed as:
\begin{equation}
	\label{eq:alg1}
	W_{k+1} = 
	\begin{cases}
		W_k - \gamma_k G(W_k;B_k), \;\;\;\;\text{if}\;p\;\text{does not divide}\;k \\
		\myparam{W_k - \gamma_k G(W_k;B_k)}A_n, \;\;\;\;\text{if}\;p\;\text{divides}\;k
	\end{cases} 
\end{equation} 
where 
$ G(W_k;B_k)=[\nabla \widetilde{f_1}(w_{k,1};B_{k,1}), \ldots, \nabla \widetilde{f_n}(w_{k,n};B_{k,n})]$
are the stochastic gradients on $n$ nodes  computed with local mini-batches.
 $A_n$ is a $n$ by $n$ matrix where every value is  $1/n$ -- it averages the model parameters on all nodes.

 \textbf{Convergence Rate.}
 As we described in \S\ref{sec:back2}, the analysis for SGD on non-convex optimization commonly uses  the weighted average of the squared gradient norms over iterations, i.e.,
 \vspace{-0.5em}
 \begin{equation}
\label{eq:metric1}
	\mathbb{E} \myrect{\sum_{k=0}^{K-1} \frac{\gamma_k}{\sum_{j=0}^{K-1}{\gamma_j}} \norm{\nabla f\myparam{w_k}}}
\end{equation}
  as the metric of convergence. 
  An algorithm is considered to have better convergence if it has smaller value of (\ref{eq:metric1}).  

For Algorithm~\ref{alg1}, we let $w_k=W_k\dotavg{n}$ (i.e., the average of model parameters on all nodes). 
Suppose the objective function is Lipschitz smooth with constant $L$ and the variance of stochastic gradient computed with any sample has an upper bound (These assumptions are commonly used in analysis for convex and non-convex optimization~\cite{doi:10.1137/070704277, doi:10.1137/120880811}). Then,  
one can show that
 \vspace{-0.3em}
\begin{flalign}
\label{eq:step8}
\mathbb{E}\norm{\nabla f\myparam{W_k\dotavg{n}}}& \leq 
 \frac{2}{\gamma_k}\myparam{\mathbb{E}f\myparam{W_{k}\dotavg{n}} - \mathbb{E}f\myparam{W_{k+1}\dotavg{n}}}\nonumber\\
	& +L^2\mathbb{E}\myrect{\Var{W_k}} +\frac{L\gamma_k\sigma^2}{M}
\end{flalign}
where 
\begin{equation}
	\mathbb{E}\myrect{\Var{W_k}} \triangleq \mathbb{E}\myrect{\frac{1}{n}\sum_{i=1}^{n}{\norm{W_k\dotavg{n} - w_{k,i}}}}
\end{equation} 
is the variance of model parameters among $n$ nodes in iteration $k$. (Readers are referred to~\cite{NIPS2018_7519} for more details.)
Summing up (\ref{eq:step8}) over $K$ iterations with weight $\gamma_k$, we can obtain
\begin{flalign}
\label{eq:bound1}
\mathbb{E} &\myrect{\sum_{k=0}^{K-1} \frac{\gamma_k}{\sum_{j=0}^{K-1}{\gamma_j}} \norm{\nabla f\myparam{W_k\dotavg{n}}}}
 \leq 
 \frac{2\myparam{f(w_0)-f(w^*)}}{\sum_{j=0}^{K-1}\gamma_k} + \nonumber\\
	& L^2\mathbb{E} \myrect{\sum_{k=0}^{K-1} \frac{\gamma_k\Var{W_k}}{\sum_{j=0}^{K-1}{\gamma_j}} }+\frac{\sum_{j=0}^{K-1}\gamma_k^2}{\sum_{j=0}^{K-1}\gamma_k}\cdot \frac{L\sigma^2}{M}
\end{flalign}
This indicates that a smaller weighted average of $\Var{W_k}$, i.e., 
\begin{equation}
\label{eq:metric2}
	\mathbb{E} \myrect{\sum_{k=0}^{K-1} \frac{\gamma_k\Var{W_k}}{\sum_{j=0}^{K-1}{\gamma_j}} }
\end{equation}
leads to smaller value of (\ref{eq:metric1}) and thus better convergence of periodic averaging SGD. 
Note that the above discussion not only applies to constant averaging period but also to other periodic averaging strategies.  
Intuitively, the algorithm has good convergence if the trajectories of model parameters on different nodes are not far apart. 

To complete the proof for CPSGD, one can show that
\begin{flalign}
\label{eq:step10}
	\mathbb{E} &\myrect{\sum_{k=0}^{K-1} \frac{\gamma_k\Var{W_k}}{\sum_{j=0}^{K-1}{\gamma_j}} } \leq    \frac{\gamma^2npC_1}{1-3\gamma^2np^2L^2} + \nonumber \\
	& \frac{3\gamma^2np^2}{1-3\gamma^2np^2L^2}\cdot\frac{1}{K}\sum_{k=0}^{K-1}\mathbb{E}\norm{\nabla f\myparam{W_{k}\dotavg{n}}}
\end{flalign}
where $p>1$ is the averaging period and $C_1$ is a constant that depends on the variance of stochastic gradients. (See~\cite{NIPS2018_7519} for more details.)
Plugging (\ref{eq:step10}) into (\ref{eq:bound1}) will show the algorithm converges at rate $O(1/\sqrt{MK})$ if $\gamma_k=\gamma \propto \sqrt{M/K}$.  

The bound in (\ref{eq:step10}) suggests that a larger averaging period $p$ leads to a larger  value of (\ref{eq:metric2}) and thus slower convergence of the algorithm.  
However, a larger averaging period also means less communication and thus faster execution of the program. 
Therefore, given the same amount of communication, we want to minimize the the value of (\ref{eq:metric2}). 
In the next section, we will show that constant averaging period is not a necessary condition for achieving $O(1/\sqrt{MK})$ convergence rate of periodic averaging SGD, and an adaptive scheduling of the averaging period will achieve smaller value of (\ref{eq:metric2}) with the same or even smaller amount of communication overhead.


%% file: text/algorithm.tex
\section{Adaptive Periodic Averaging SGD}

In this section, we describe our main idea of using adaptive averaging period to achieve a good trade-off between communication and convergence for distributed SGD. 

\subsection{Inefficiency of Constant Periodic Parameter Averaging SGD}

To motivate our algorithm, we first illustrate the problem of  Constant Periodic parameter averaging SGD (CPSGD) in Algorithm~\ref{alg1}. Suppose the learning rate  over the $p$ iterations between two synchronization steps in Algorithm~\ref{alg1} is constant, we define the average variance of the model parameters between two synchronization steps  as 
\begin{equation}
\label{eq:vt}
	V_t \triangleq \frac{1}{p}\sum_{k=pt}^{p(t+1)-1}{\Var{W_k}}.
\end{equation}
If we view  $p$ iterations as a complete training process, the bound in (\ref{eq:step10}) suggests that $V_t$ is proportional to ${\gamma}_{pt}^2$.    
This indicates that $V_t$ will decrease if the learning rate diminishes during the training process.         
Also, because the second term on the right hand side of (\ref{eq:step10}) (i.e., the squared gradient norm)  diminishes over iterations,  $V_t$ is expected to decrease even if the learning rate does not change. 

\begin{figure}
	\centering
	\includegraphics[scale=0.39]{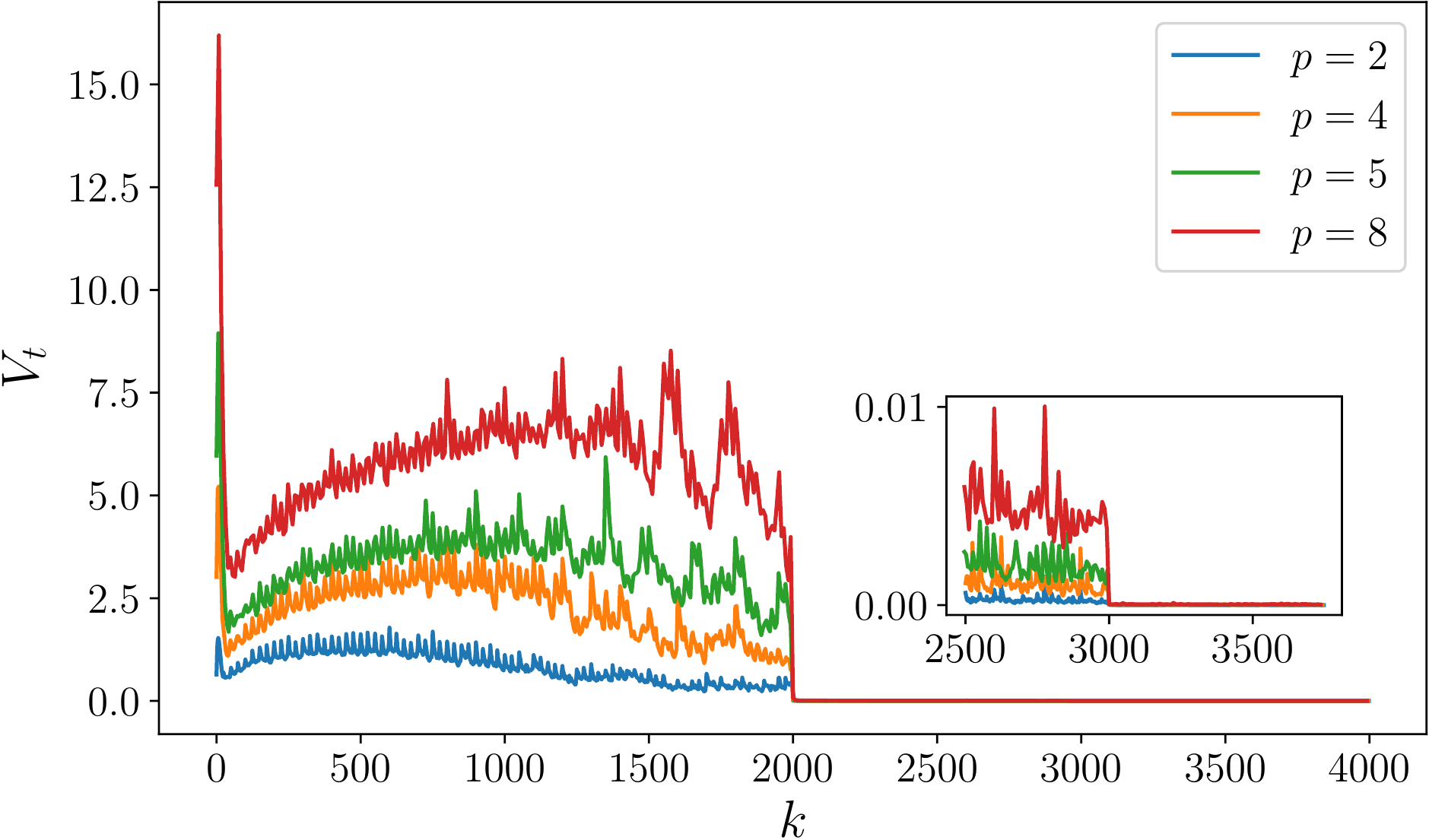}
	\vspace{-0.5em}
	\caption{Average variance ($V_t$) of model parameters  between two synchronizations (across iterations, $k$)   in Constant Period SGD ($p$ is period for synchronization)  Training GoogLeNet with CIFAR-10 dataset on 16 nodes.}
	\label{fig:googlenetvars}
\end{figure} 


To validate this point, we conduct an experiment by training GoogLeNet~\cite{DBLP:journals/corr/SzegedyLJSRAEVR14} on CIFAR-10 dataset~\cite{Krizhevsky09learningmultiple} with CPSGD on 16 nodes.   
The learning rate is initialized to $0.1$ and annealed to $0.01$ and $0.001$ after $80$ and $120$ epochs, respectively. 
The total number of epochs is $160$. 
The mini-batch size on each node is set to $128$, so the total mini-batch size is $128\times 16=2048$.
We test four constant averaging period values: $2$, $4$, $5$,  and $8$. 
Figure~\ref{fig:googlenetvars} shows  the values of $V_t$ over iterations ($k$) with the four averaging periods. 
We can see that the variance is extremely large initially, and drops to a small number after the first few  
iterations. 
Then, the variance increases a bit and then again starts decreasing. 
Also, the  variance value drops after the $80$th and $120$th epochs -- recall that the learning rate is decreased by a factor of $10$ in each of these cases. 
This experimental result is  consistent with the above discussion that {\em $V_t$ is proportional to $\gamma_{pt}^2$ and decreases with the magnitude of the gradient in CPSGD}. 

Let us consider applying the following four averaging strategies to the above training process:
\begin{enumerate}[\;\itshape 1.]
	\item The averaging period is $4$ for the first $80$ epochs and increases  to $8$ for the remaining epochs;
	\item The averaging period is $8$ for the first $80$ epochs and decreases to $4$ for the remaining epochs;
	\item The averaging period is $8$ for all $160$ epochs;
    \item The averaging period is $5$ for all $160$ epochs.
\end{enumerate} 
It should be noted that  strategy-$1$ and strategy-$2$ have the same communication overhead, as the total number of synchronization steps  in both cases is $750$ ($2000/4+2000/8$). 
Because strategy-$1$ uses communication period $4$ in the first $2000$ iterations,  it corresponds to the line of $p=4$ in Figure~\ref{fig:googlenetvars} in the first 2000 iterations. Similarly, strategy-$2$ and strategy-$3$ corresponds to the line of $p=8$ in the first $2000$ iterations. 
Therefore, in terms of convergence,  comparing strategy-$1$ and strategy-$2$ with strategy-$3$,  we can see that strategy-$1$ reduces $V_t$ for $k \in [0, 1999]$ by about $2.5$ according to Figure~\ref{fig:googlenetvars}, while strategy-$2$ reduces $V_t$ for $k \in [2000, 3999]$ by about $0.005$.    
Because 
\begin{equation}
\label{eq:equals}
	\mathbb{E} \myrect{\sum_{k=0}^{K-1} \frac{\gamma_k\Var{W_k}}{\sum_{j=0}^{K-1}{\gamma_j}} }=\mathbb{E} \myrect{\sum_{t=0}^{T-1} \frac{\gamma_{pt} V_t}{\sum_{j=0}^{T-1}{\gamma_{pj}}} }
\end{equation}
where $T=K/p$, strategy-$1$  achieves a much smaller value of (\ref{eq:metric2}) than strategy-$2$.  
Since a smaller value of (\ref{eq:metric2}) indicates better convergence, strategy-$1$ converges faster than  strategy-$2$ with the same communication overhead.

Comparing strategy-$1$ with strategy-$4$, we can see that strategy-$1$ reduces $V_t$ for $k \in [0, 1999]$ by about $2$, and increases $V_t$ for $k \in [2000, 3999]$ by less than $0.01$.  
Because the total number of synchronizations in strategy-$4$ is $800$ ($4000/5$),  
strategy-$1$ achieves better convergence than strategy-$4$ while requiring even less communication. 
  In other words, if we want to achieve the convergence of strategy-$1$ with constant periodic averaging, we must use a period less than $5$ (i.e., more than $800$ synchronization steps).  

The above example can also be explained with the bound in (\ref{eq:step10}). 
At the beginning of the training process when the learning rate  and the gradient are large, reducing $p$ leads to significant decrease of the right hand side of (\ref	{eq:step10}). 
In contrast,   
when the learning rate and the gradient are small in later phase of the training process, reducing $p$ will bring only small  benefits because the right hand side of (\ref	{eq:step10}) is already small.

\subsection{Increasing Averaging Period Adaptively}
\label{sec:adpsgd}
We have shown that to minimize the value of (\ref{eq:metric2}) for periodic averaging SGD with the same communication overhead, one should use small averaging period at first and gradually increase the period during the training process. 
The question now is how to determine the proper frequency and magnitude of the increase so that the communication overhead can be reduced without sacrificing convergence rate to  a large 
degree. 
We now introduce an adaptive period scheduling strategy to address this problem. 
 
According to the bound in~(\ref{eq:step8}),  
if we can ensure
\begin{flalign}
	\label{eq:step9}
	\mathbb{E}\myrect{\Var{W_k}}\leq  \gamma_k\frac{C_2}{M}
\end{flalign} 
where $C_2$ is a constant, the second term on the right hand side of (\ref{eq:bound1}) will be
\begin{flalign}
	\label{eq:bound2}
	L^2\mathbb{E} \myrect{\sum_{k=0}^{K-1} \frac{\gamma_k\Var{W_k}}{\sum_{j=0}^{K-1}{\gamma_j}} } \leq \frac{\sum_{j=0}^{K-1}\gamma_k^2}{\sum_{j=0}^{K-1}\gamma_k}\cdot \frac{L^2C_2}{M}
\end{flalign} 
which is bound to $O(1/\sqrt{MK})$ with a proper scheduling of $\gamma_k \propto \sqrt{M/K}$. 
Because the first and the third term on the right hand side of (\ref{eq:bound1}) are both bound to $O(1/\sqrt{MK})$ with a proper configuration of $\gamma_k$, 
the algorithm will achieve $O(1/\sqrt{MK})$ convergence rate if condition (\ref{eq:step9}) holds. 
According to (\ref{eq:vt}), condition (\ref{eq:step9}) implies
\begin{equation}
\label{eq:vtpt}
	V_t \leq \gamma_{pt}\frac{C_2}{M}.
\end{equation}
Recall that CPSGD has $V_t$ proportional to $\gamma_{pt}^2$ and decreasing with the gradient,  (\ref{eq:vtpt}) indicates that constant averaging period is not a necessary condition for  achieving $O(1/\sqrt{MK})$ convergence rate. 
The idea of our {\em adaptive} periodic averaging is that, we can start with smaller $V_t$ (by using a smaller averaging period in early iterations) but keep $V_t$ proportional to $\gamma_{pt}$ instead of  $\gamma_{pt}^2$ (by adaptively increasing the period in later iterations).  
This strategy reduces the overall variance of model parameters while maintaining the $O(1/\sqrt{MK})$ convergence rate.  

In practice, it is infeasible  to bound $\Var{W_k}$ for all $k$'s because it requires exchanging the model parameters among nodes  in each iteration. 
Instead,  we ensure the variance of model parameters right before each synchronization step close to $\gamma_k\frac{ C_2}{M}$, i.e.,
\begin{equation} 
\label{eq:varsync}
S_k \triangleq \Var{W_{k}-\gamma G(W_{k};B_{k})} \approx \gamma_k\frac{C_2}{M} 
\end{equation} 
when synchronization happens in iteration $k$. 
The value of $S_k$ can be computed  after the averaging of model parameters with only a small overhead. 
Because $\Var{W_k}$ becomes zero after each synchronization and the variance gradually accumulates until the next synchronization, 
it is most likely that all $\Var{W_k}$'s since last synchronization step are smaller than or close to  $\gamma_k \frac{C_2}{M}$ if (\ref{eq:varsync}) holds.  
We can sample the value of $C_2$ in the first few synchronization steps  with a small averaging period. 
In later synchronization steps, if $S_k$ is smaller than $\gamma_k \frac{C_2}{M}$, we increase the averaging period; otherwise, we decrease the averaging period.

\begin{algorithm}[t]
\footnotesize
\SetAlgoLined
\Require{$p_{init}$, $K_s$}
$cnt=0$\;
$p=p_{init}$\;
$C_2=0$\;
\For{$k=0,1,2, \ldots ,K-1$}{
  $cnt += 1$\;
  $\nabla \widetilde{f}(w_{k,i};B_i) = \frac{1}{m}\sum_{j\in B_i}{\nabla F_j(w_{k,i})}$\;
  \tcc{Updating parameter variables locally}
  $w_{k+1,i}=w_{k,i}-\gamma_k \nabla \widetilde{f}(w_{k,i};B_i)$\;
  \If{$cnt == p$}{
   $cnt=0$\;
   $w_{k+1,i}'=\frac{1}{n} \sum_{j=1}^{n}{w_{k+1,j}}$\;
   \tcc{Computing the variance of model parameters among nodes}
   $S_k=\frac{1}{n}\sum_{j=1}^{n}\norm{w_{k+1,j}'-w_{k+1,j}}$\;
   $w_{k+1,i}=w_{k+1,i}'$\;
   \eIf{$k<K_s$} {
   \tcc{Sampling value of $C_2$}
   $C_2=\textproc{RunningAverage}(C_2, S_k/\gamma_k)$\;
   }
   {
   \tcc{Updating averaging period}
   \uIf{$S_k < 0.7 \times\gamma_k C_2$}{
   $p$ += $1$\;
   }
   \uElseIf{$S_k > 1.3 \times\gamma_k C_2$}{
     $p$ -= $1$\;
  }
  } 
  } 
 }
 \caption{(ADPSGD) Procedure of adaptive periodic parameter averaging SGD on the $i$th node}
 \label{alg2}
\end{algorithm}

Algorithm~\ref{alg2} shows the procedure of our adaptive periodic averaging SGD (ADPSGD). 
We use a counter $cnt$ to record the number of iterations since the last synchronization (line 1). 
The counter is incremented by $1$ in each iteration (line 5). 
The averaging period $p$ is initialized to a small value $p_{init}$ (line 2).
Once $cnt$ equals to $p$, synchronization is performed (line 8-20).  
First, $cnt$ is reset to $0$ (line 9).
Next, the model parameters are averaged among the nodes (line 10). 
Then, $S_k$ is computed by averaging the squared deviation of the model parameters on all nodes (line 11).   
The computation of $S_k$ requires one more synchronization, but the data transferred is a single  
floating-point value, and  thus, it incurs only a very small  overhead.  
When the iterate number $k$ is smaller than a given threshold $K_s$, we compute the running average of $S_k/\gamma_k$, which will be used as the sampled  $C_2$ for the rest of the iterations (line 14). 
After the sampling phase, we compare  $S_k$ with $\gamma_k C_2$.
If $S_k$ is smaller than $0.7\times\gamma_kC_2$, we increase the averaging period by $1$ (line 17). 
If $S_k$ is larger than $1.3\times\gamma_kC_2$, we decrease the averaging period by $1$ (line 19). 
We use $0.7$ and $1.3$ in line 16 and 18 because we need values  slightly smaller than 1 and slight greater than 1, respectively  to keep $S_k$ close to $\gamma_k\frac{C_2}{M}$,

\begin{figure}
	\centering
	\includegraphics[scale=0.38]{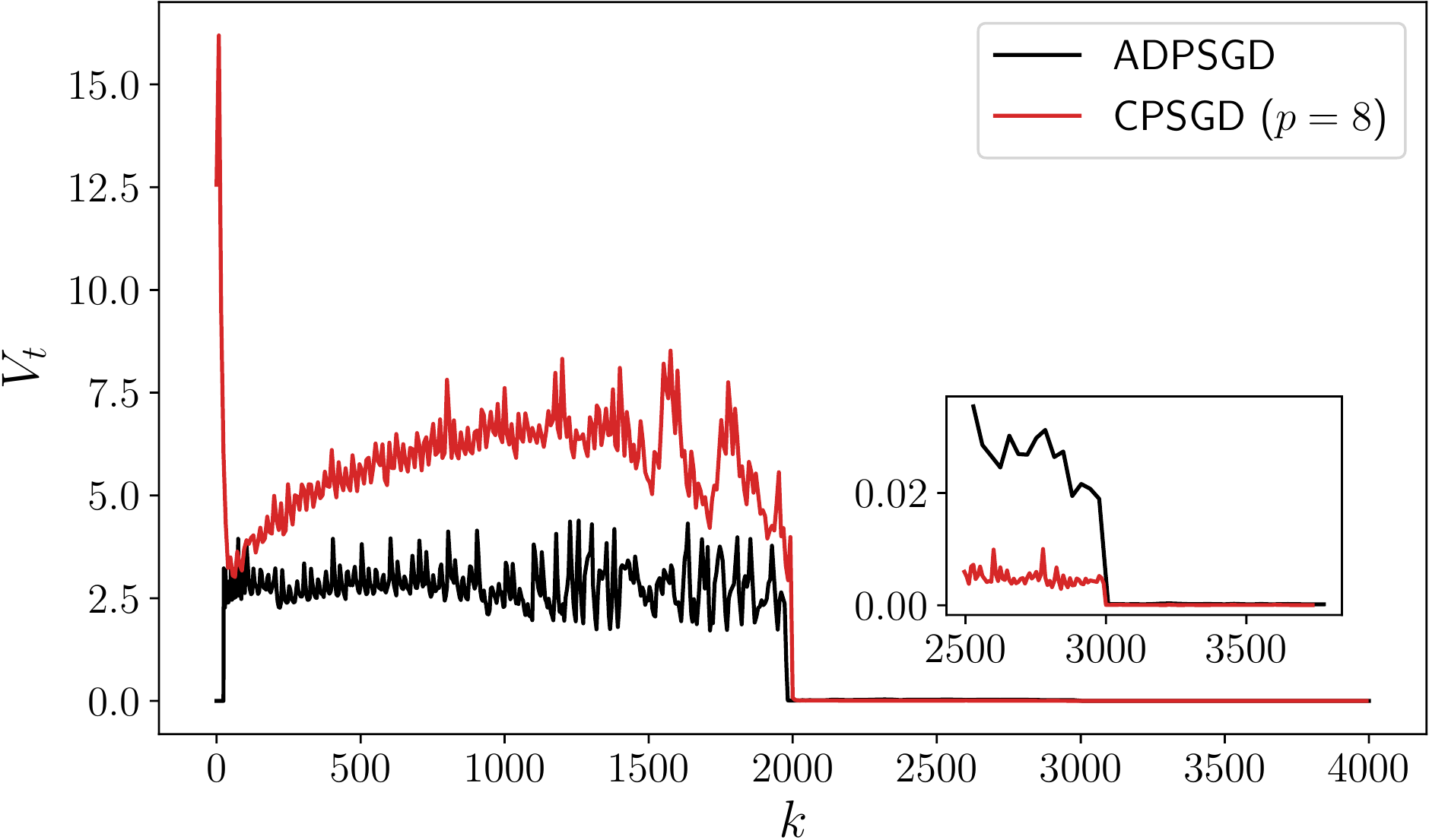}
	\vspace{-0.5em}
	\caption{Average variance of model parameters between two synchronizations in ADPSGD  for training GoogLeNet with CIFAR-10 dataset on 16 nodes.}
	\label{fig:googlenetadpvars}	
\end{figure}

To show how our ADPSGD changes $\Var{W_k}$, we apply ADPSGD to train the same model as in Figure~\ref{fig:googlenetvars}. 
Figure~\ref{fig:googlenetadpvars} shows $V_t$ of our ADPSGD. 
We also plot $V_t$ of CPSGD with $p=8$ in Figure~\ref{fig:googlenetadpvars} for comparison.   
With  ADPSGD, the first epoch uses an  averaging period of $1$ to avoid the large initial variance. 
Then, we apply Algorithm~\ref{alg2} with $p_{init}=4$ and $K_s=1000$. 
We can see that  $V_t$ is almost constant in ADPSGD in the first $2000$ iterations. 
From the zoomed-out portion of the   figure, we can see that ADPSGD has a larger $V_t$ than CPSGD in later  
phase of the training. 
This experimental result validates our discussion in this section that our ADPSGD starts with smaller $V_t$ and maintains a slower decrease of $V_t$ compared with constant periodic averaging.  
According to (\ref{eq:equals}), it is obvious from Figure~\ref{fig:googlenetadpvars} that our ADPSGD has a much smaller value of (\ref{eq:metric2}) (and thus better convergence) than CPSGD with $p=8$. 

\begin{figure}
	\centering
	\includegraphics[scale=0.38]{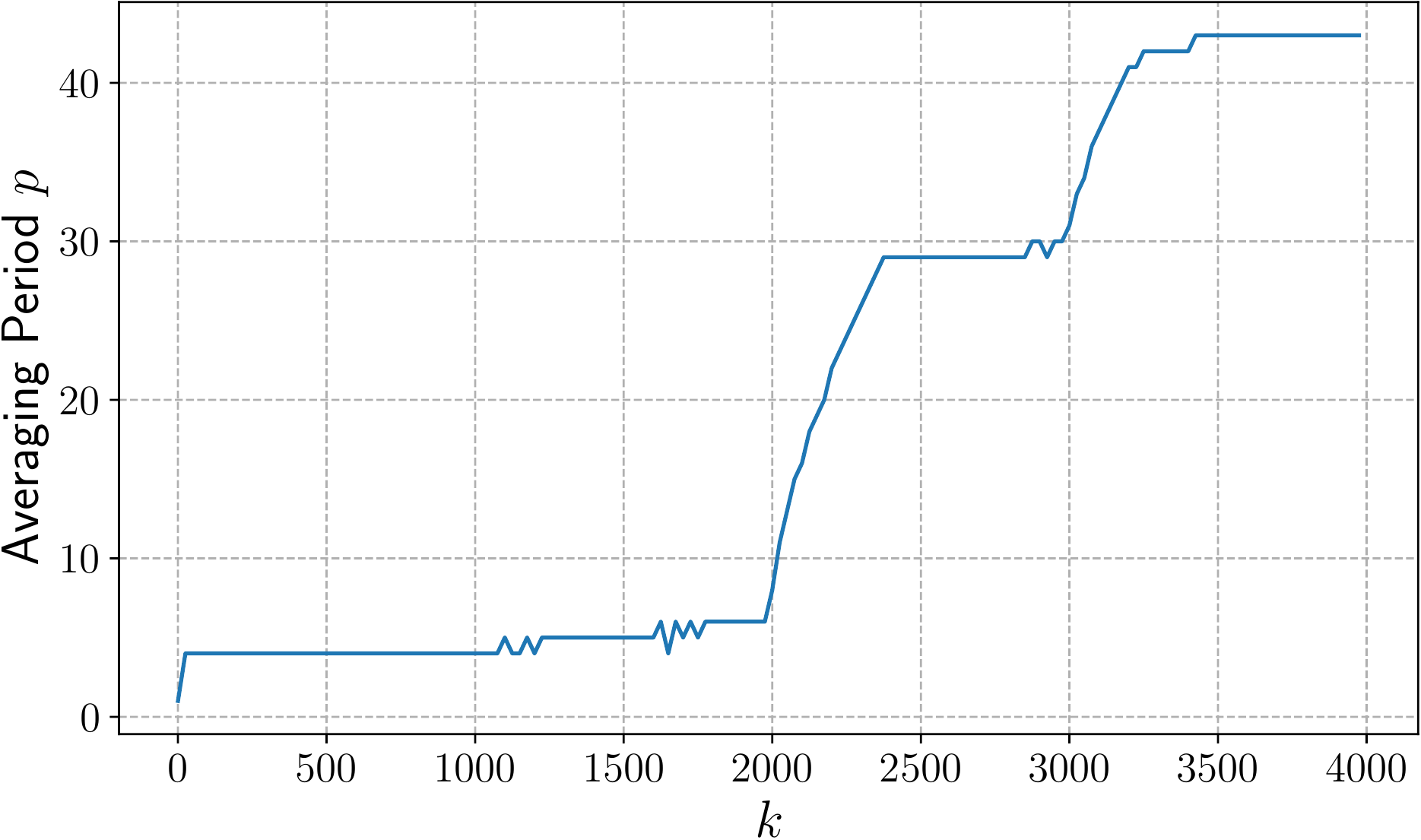}
	\vspace{-0.5em}
	\caption{Averaging Period in ADPSGD  for training GoogLeNet with CIFAR-10 dataset on 16 nodes.}
	\label{fig:googlenetadpp}	
\end{figure}

Figure~\ref{fig:googlenetadpp} shows the averaging period in ADPSGD over the training process. 
The averaging period is fixed to $4$ in the first $1000$ iterations for sampling the scaler $C_2$ in Algorithm~\ref{alg2}. 
After the sampling, the algorithm adjusts the averaging period automatically. 
We can see that the averaging period gradually increases to $6$ in the first $2000$ iterations. 
Starting from iteration $2000$ when the learning rate is decreased to $0.01$, the averaging period gradually increases to $29$. 
After iteration $3000$ when the learning rate is decreased to $0.001$, the averaging period further increases to $43$. 
The total number of synchronization steps is $498$, so the communication overhead is close to CPSGD with $p=4000/498\approx8.03$. 

 Figure~\ref{fig:googlenetadpvars} and~\ref{fig:googlenetadpp} show that our ADPSGD achieves better convergence than CPSGD with even less communication. 
 One may suspect the large averaging period in later phase of the training process could cause slow convergence. 
However, note that the condition in (\ref{eq:step9}) holds throughout the training process and the convergence rate of ADPSGD is guaranteed as we discussed in this section. 
As we will show through more detailed  experiments, our ADPSGD indeed achieves smaller training loss and  
higher test accuracy than CPSGD for training different neural networks on different datasets. 
In fact, our ADPSGD even achieves higher or equal test accuracy compared with vanilla mini-batch SGD (i.e., CPSGD with $p=1$) for all test cases in our experiments. 
This is because periodic averaging is  helpful for large-batch SGD to escape sharp minima and avoid overfitting, especially for training with large batches. 
We explain this  in Section~\ref{sec:twlbs}.

%% file: text/experiments.tex
\section{Evaluation}

\begin{figure*}
	\centering
	\subfloat[Training Loss]{
	\includegraphics[scale=0.33]{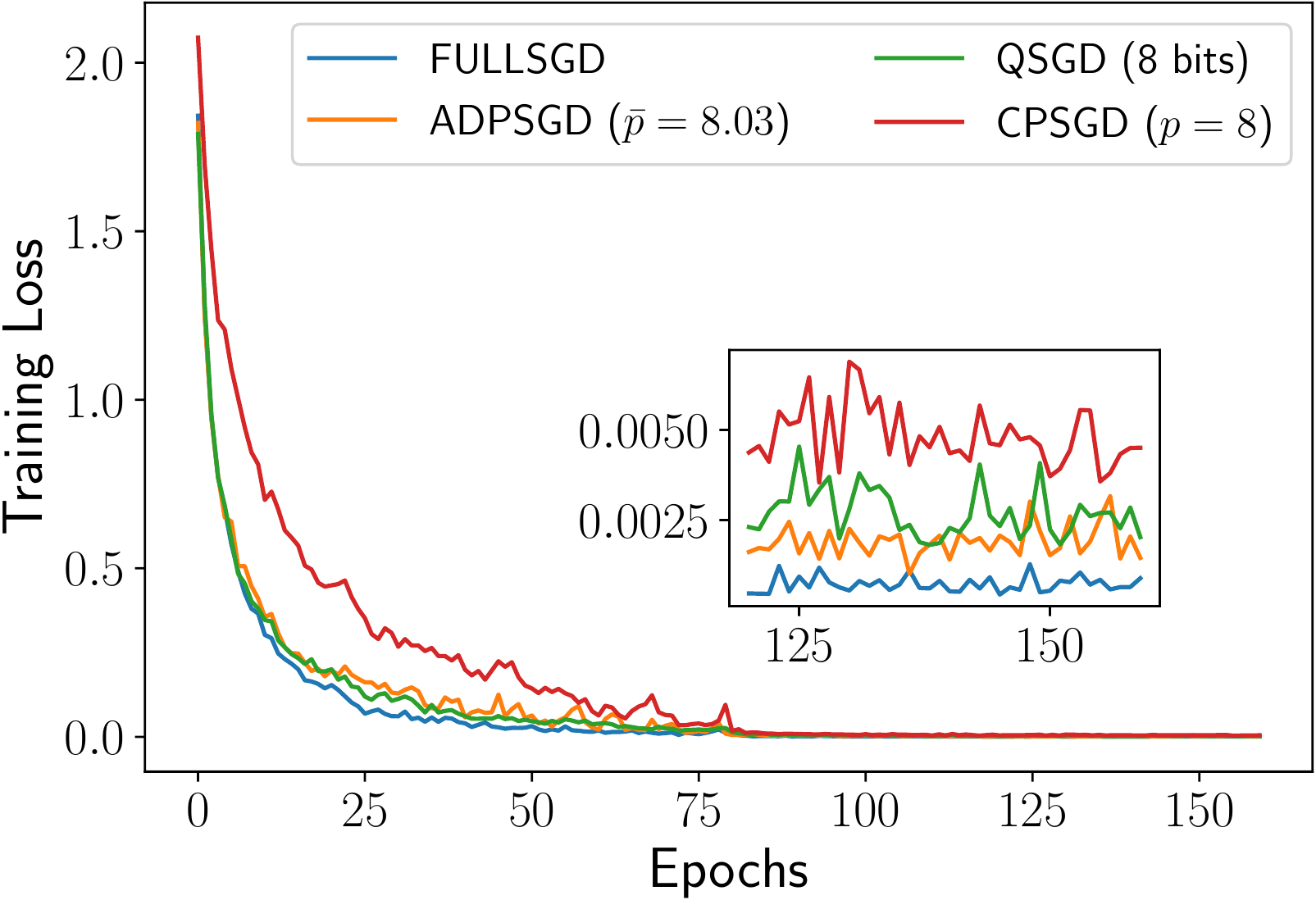}
	\label{fig:googlenetloss}
	}
	\subfloat[Test Accuracy]{
	\includegraphics[scale=0.33]{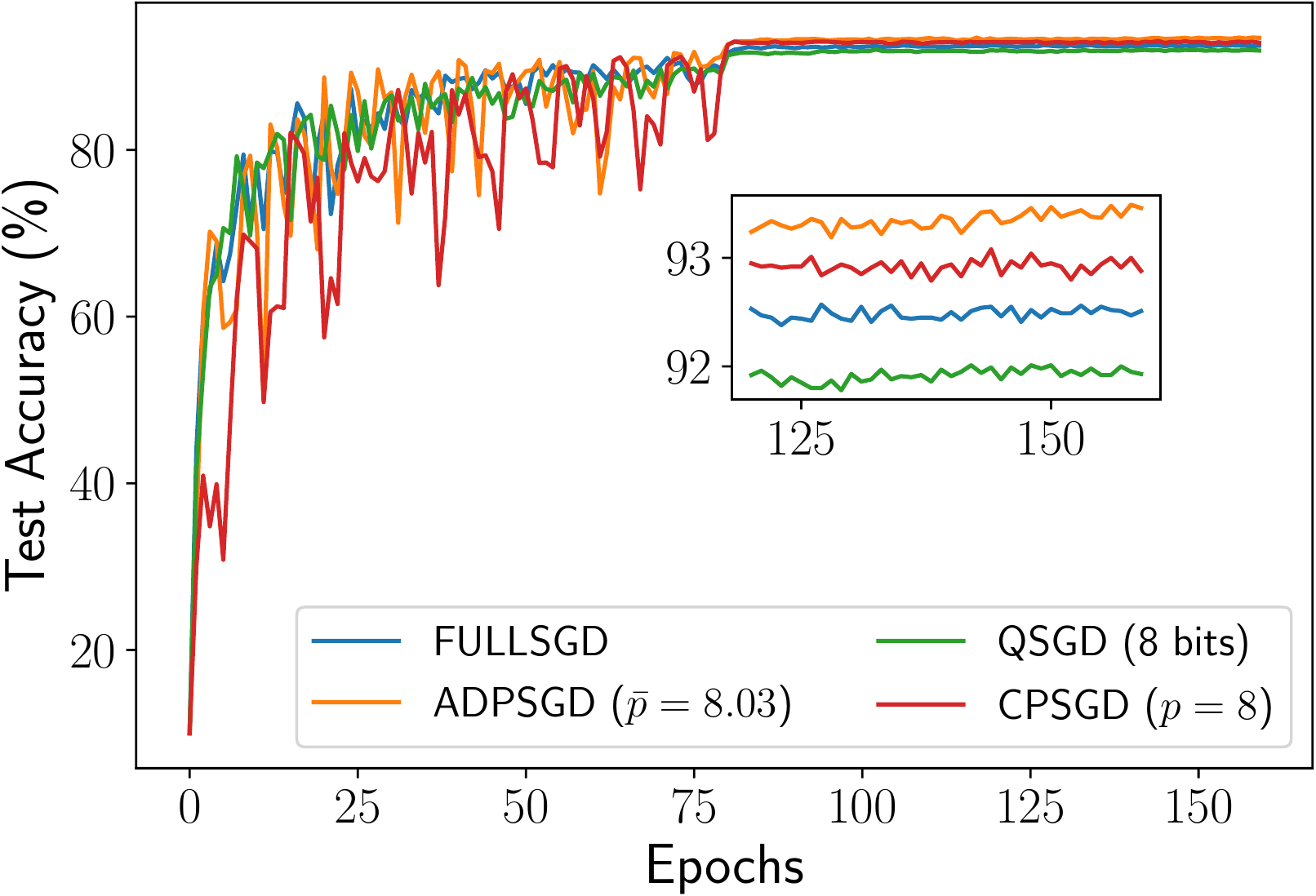}
		\label{fig:googlenetacc}
	}
	\subfloat[Execution Time]{
	\includegraphics[scale=0.34]{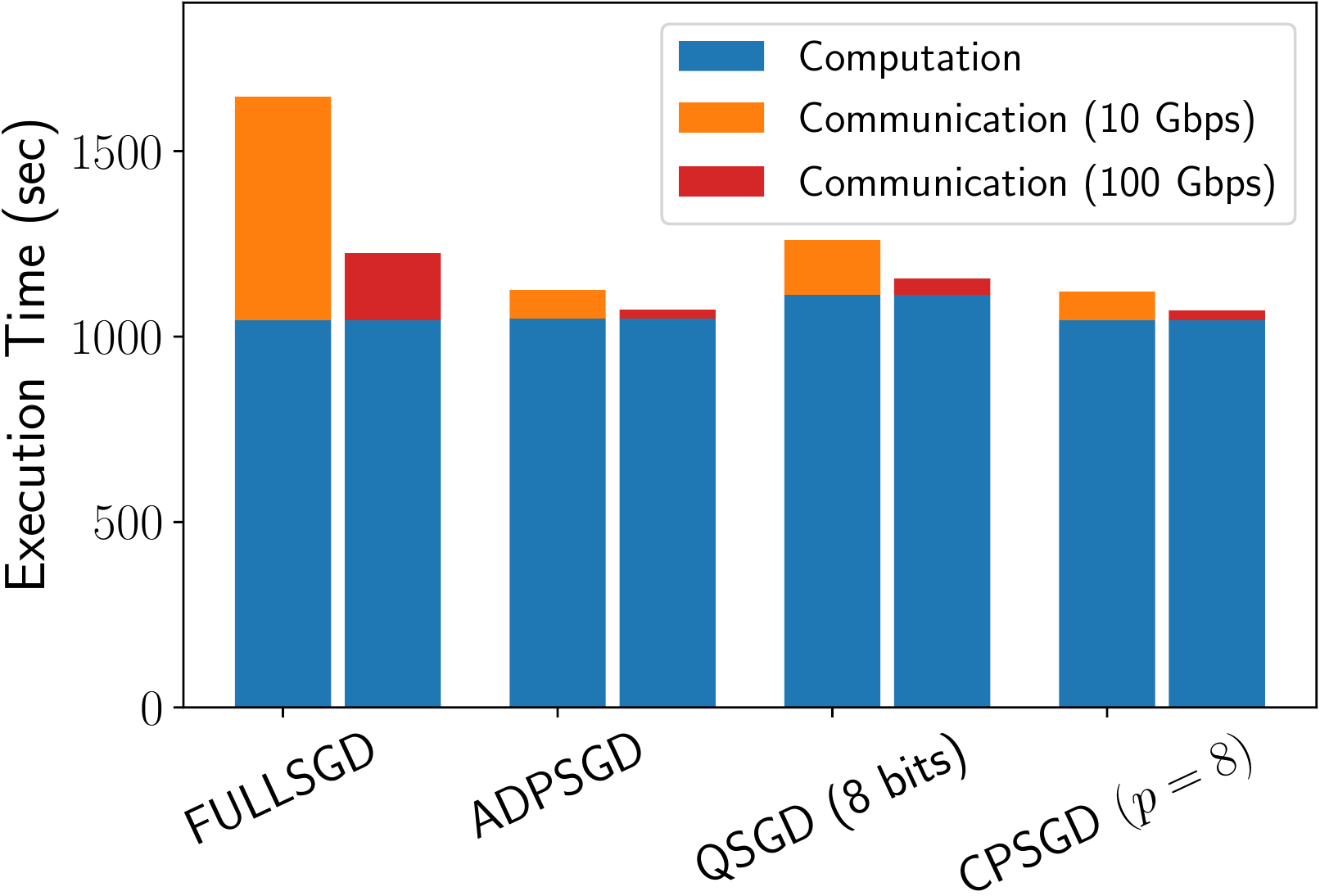}
		\label{fig:googlenettime}
	}
	\vspace{-0.5em}
	\caption{Training GoogLeNet on CIFAR-10 with 16 GPUs}
	\label{fig:googlenet}
\end{figure*}

\begin{figure*}
	\centering
	\subfloat[Training Loss]{
	\includegraphics[scale=0.33]{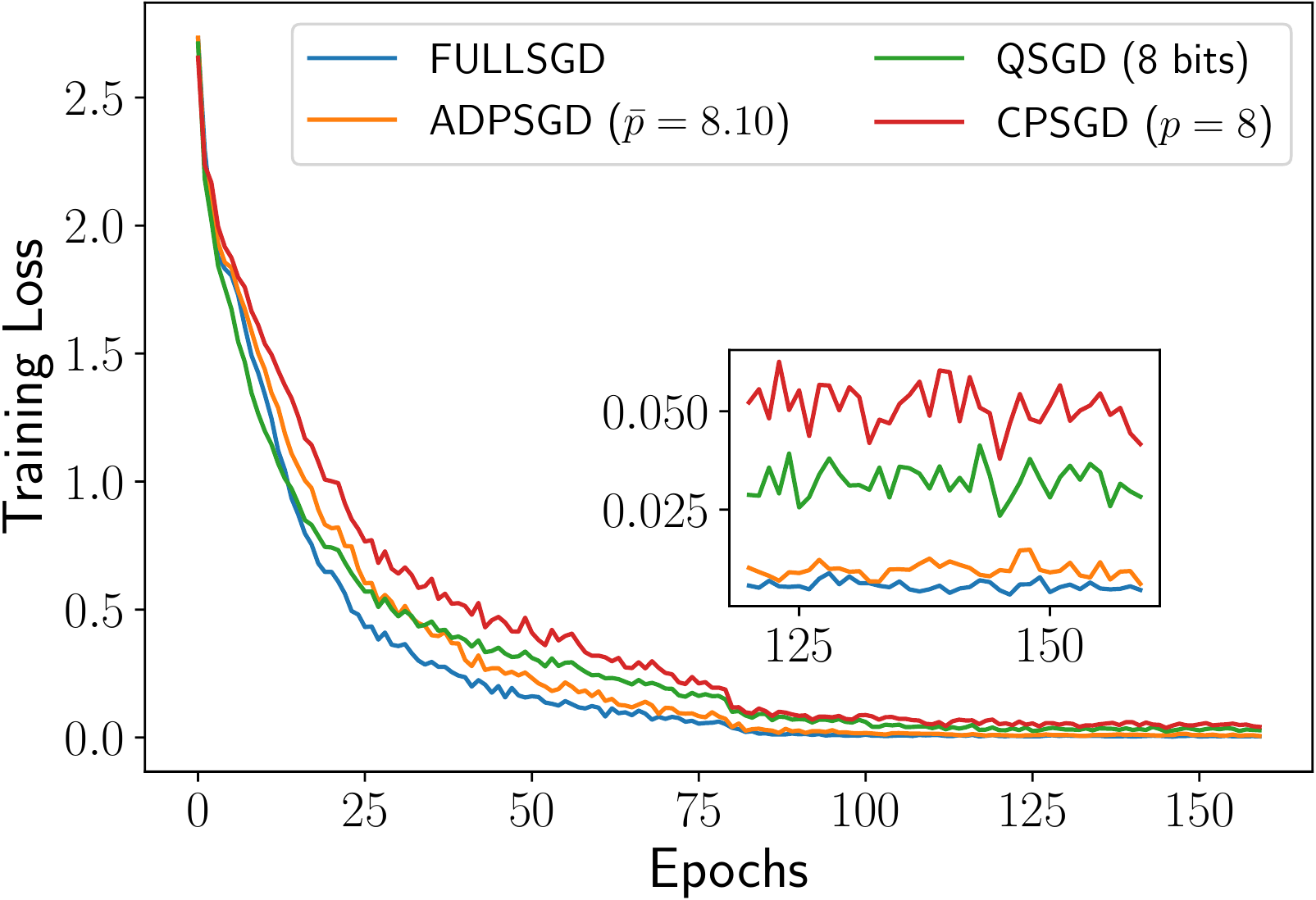}
	\label{fig:vggloss}
	}
	\subfloat[Test Accuracy]{
	\includegraphics[scale=0.33]{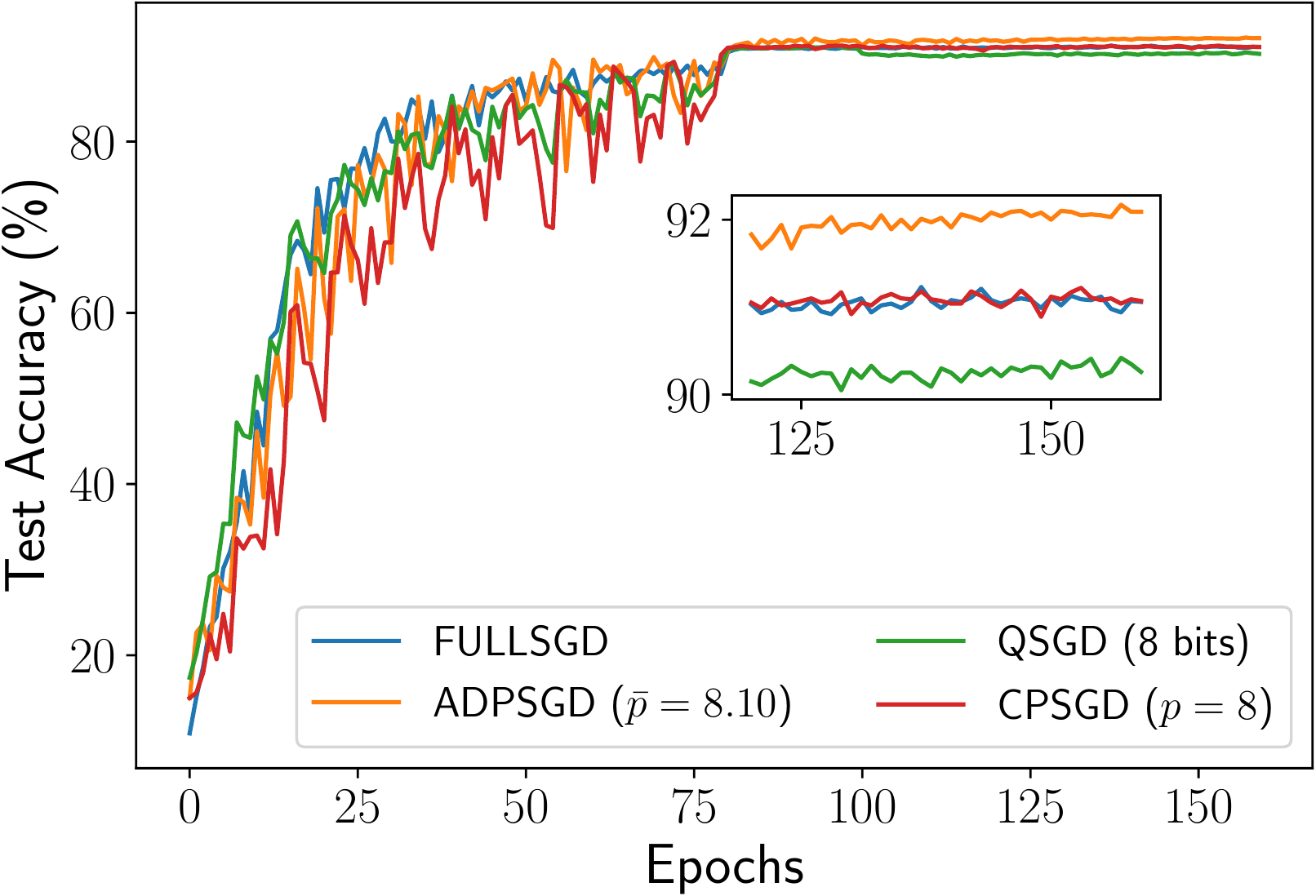}
	\label{fig:vggacc}
	}
	\subfloat[Execution Time]{
	\includegraphics[scale=0.34]{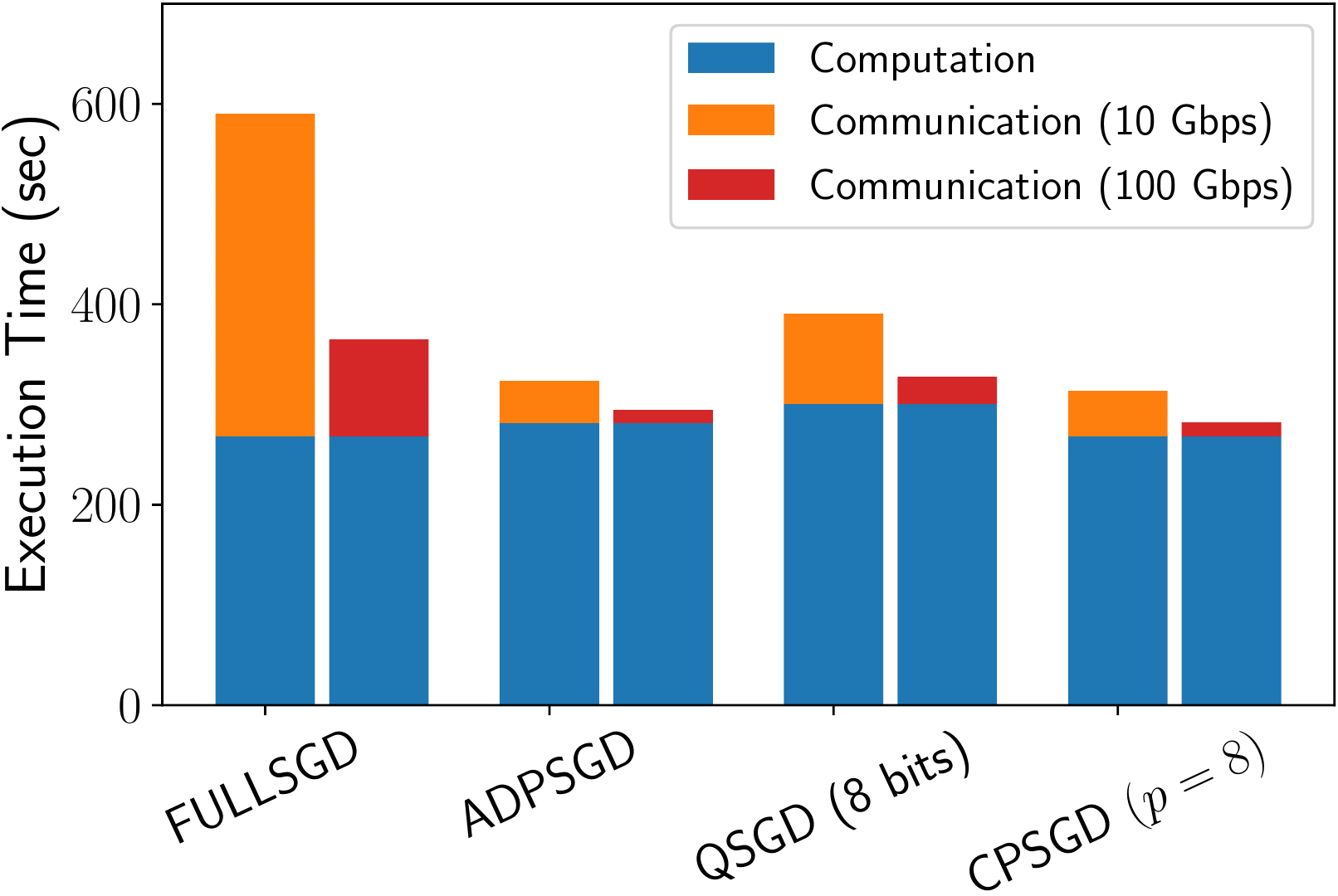}
			\label{fig:vggtime}
	}
	\vspace{-0.5em}
	\caption{Training VGG16 on CIFAR-10 with 16 GPUs}
		\vspace{-1em}
		\label{fig:vgg}
\end{figure*}

In this section, we evaluate our ADPSGD and compare its performance with full-communication SGD, CPSGD, and gradient-quantization SGD  on several image classification benchmarks, using a GPU cluster.  

\subsection{Experimental Setup}

We train GoogLeNet and  VGG16~\cite{DBLP:journals/corr/SimonyanZ14a} on CIFAR-10 dataset, and \revision{ train ResNet50 and AlexNet on  ILSVRC 2012 (ImageNet)}. 
The experiments are conducted on 16 nodes each equipped with an Nvidia Tesla P100 GPU. 
The nodes are connected with 100Gbps InfiniBand based on a fat-tree topology. 
GPUDirect peer-to-peer communication is supported. 
We implement the algorithms in the paper within PyTorch 1.0.0.
We use NCCL 2.3.7 for CUDA 9.2 as the communication backend.  
The training data is stored in a shared file system, and are globally shuffled at the end of each epoch. 
To achieve a good utilization of the GPUs, we use mini-batch size of $128$ on each node for all test cases. 
The total mini-batch size on $16$ nodes is $128\times 16=2048$. 

We compare four versions of SGD. 
{\tt FULLSGD} is the vanilla mini-batch SGD with full communication. 
{\tt CPSGD} is constant periodic averaging SGD (Algorithm~\ref{alg1}) with a communication period of 8. 
{\tt ADPSGD} is our proposed adaptive periodic parameter averaging SGD (Algorithm~\ref{alg2}). 
{\tt QSGD} is gradient-quantization SGD proposed in~\cite{NIPS2017_6768}. 
For {\tt QSGD}, we use 8 bits to store the quantized value for each gradient component. 
For all versions, we set momentum coefficient to 0.9, which is common used for training CNN models~\cite{DBLP:journals/corr/abs-1708-01012, NIPS2017_7117}. 

\revision{All tests were run five times. Because the differences of final training loss and accuracy across executions were small, we report only one set of the results.}

\subsection{Results on CIFAR-10}
 
We set the initial learning rate to $0.1$. 
The learning rate is annealed to $0.01$ and $0.001$ at epoch $80$ and epoch $120$, respectively. 
The total number of epochs is $160$. 
For ADPSGD, we use averaging period of $1$ for the first epoch for warmup, and then apply Algorithm~\ref{alg2} with $p_{init}=4$, $K_s=0.25K$ where $K$ is the total number of training iterations. 
Though we report results with this specific configuration of Algorithm~\ref{alg2}, the performance of our ADPSGD is not sensitive to and thus does not require fine tuning of $p_{init}$ and $K_s$. 
In fact, we achieve almost the same final test accuracy with $p_{init}$ from $2$ to $5$ and $K_s$ from $500$ to $1500$. 
When $p_{init}$ is set to $8$, the best accuracy of ADPSGD decreases $0.5\% \sim 1.0\%$.


\begin{table}[t]
\footnotesize
\centering
\begin{tabular}{c|c|c|c|c}
\hline
\hline
 & {\tt SMALL\_BATCH} & {\tt ADPSGD} & {\tt CPSGD} & {\tt FULLSGD} \\
 \hline
 GoogLeNet & $93.68\%$ & $93.49\%$ & $93.08\%$ & $92.94\%$   \\
 \hline
 VGG16 & $92.45\%$ & $92.17\%$ & $91.29\%$ & $92.10\%$  \\
\hline
\end{tabular}
\caption{Best test accuracy on CIFAR-10 achieved by (a) {\tt SMALL\_BATCH}: vanilla SGD  with batch size $128$ and initial learning rate $\gamma_0=0.1$;  (b) ADPSGD with  batch size $2048$ and $\gamma_0=0.1$; (c) CPSGD with batch size $2048$, $\gamma_0=0.1$ and averaging period $p=2,3,\ldots,16$; (d) FULLSGD with batch size $2048$ and $\gamma_0=0.1,0.2,\ldots,1.6$.}
\label{tab:accuracies}
\end{table}

Comparing our ADPSGD with CPSGD  in Figures~\ref{fig:googlenet} and~\ref{fig:vgg}, we can see that our ADPSGD achieves smaller/equal training loss and higher test accuracy. 
The total number of synchronization steps in ADPSGD is $498$ and $494$ for training GoogLeNet and VGG16, 
 respectively.  
Thus, the communication overhead of ADPSGD is close to CPSGD with $p=4000\div 498\approx 8.03$ and $p=4000/494\approx 8.1$ in the two cases.  
This means our ADPSGD  requires slightly less overall communication than CPSGD with $p=8$. 
To further validate the advantage of ADPSGD over CPSGD, we test CPSGD with averaging period from $2$ to $16$. 
The best accuracy in each case with CPSGD is shown in the third column of Table~\ref{tab:accuracies}. 
For GoogLeNet, the best accuracy of CPSGD is achieved with $p=7$, and  
for VGG16, the best accuracy of CPSGD is achieved  with $p=4$ -- in both cases, this will require more commmunication
as compared to ADPSGD. 
Furthermore, the accuracy achieved is still lower  than the accuracy levels  achieved by our ADPSGD (the second column of Table~\ref{tab:accuracies}). 
The results validate our discussion in \S\ref{sec:adpsgd} that our ADPSGD achieves  better convergence than CPSGD while requiring less communication. 


Comparing our ADPSGD with FULLSGD in Figure~\ref{fig:googlenet} and~\ref{fig:vgg}, we can see that our ADPSGD achieves almost the same training loss and even higher test accuracy values. 
The results of FULLSGD in Figure~\ref{fig:googlenet} and~\ref{fig:vgg}  are collected with initial learning rate of $0.1$. 
To show the advantage of our ADPSGD, we further test FULLSGD with different initial learning rates from $0.2$ to $1.6$, and compare their test accuracy with small-batch SGD (i.e., vanilla SGD with batch size $128$).     
The best test accuracy achieved by small-batch SGD and FULLSGD are shown in the first and the last column of Table~\ref{tab:accuracies}, respectively.  
We can see that small-batch SGD achieves the highest test accuracy and our ADPSGD achieves the second highest. 
For GoogLeNet, the best accuracy of FULLSGD is achieved with initial learning rate of $0.3$. 
For VGG16, the best accuracy of FULLSGD is achieved  with initial learning rate of $0.2$.  
Our ADPSGD consistently achieves higher test accuracy than different configurations of FULLSGD. 
The results validate that  our ADPSGD is effective in overcoming the generalization gap of large-batch training, whereas increasing the learning rate does not always work. 

Comparing our ADPSGD with QSGD in Figures~\ref{fig:googlenet} and~\ref{fig:vgg}, we can see that our ADPSGD achieves smaller training loss and higher test accuracy. 
Because QSGD uses 8 bits to store each gradient component, its communication data size is $1/4$ of FULLSGD and is $2$x of our ADPSGD. 
Moreover, compared to FULLSGD, our ADPSGD reduces latency in communication by a factor of $8$ while QSGD does not reduce latency. 
The results suggest that a simple periodic synchronization strategy can outperform the sophisticated gradient-compression method in terms of both convergence rate and generalization with even less amount of communication.

\begin{figure}
	\centering
	\subfloat[GoogLeNet]{
	\includegraphics[scale=0.35]{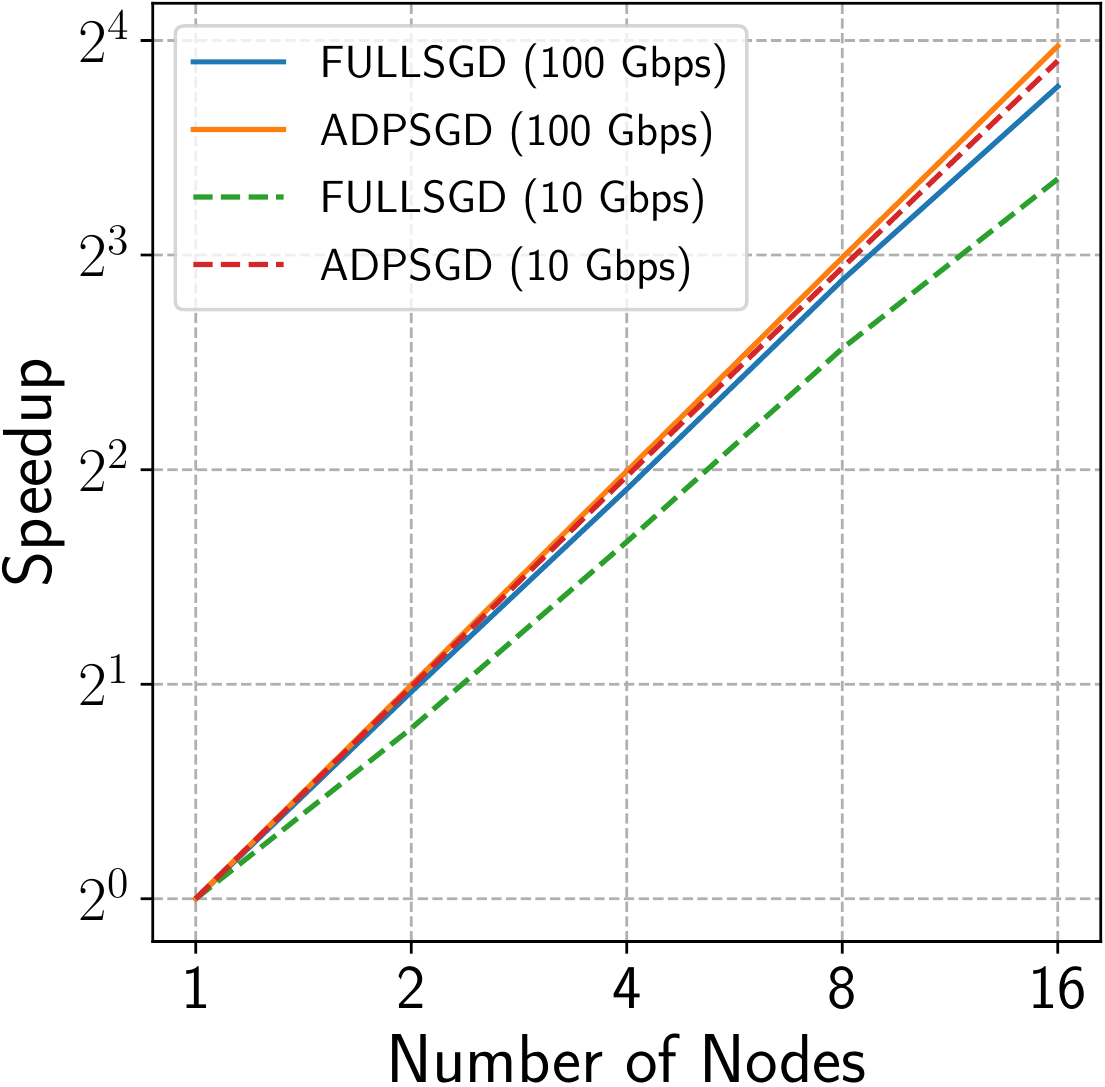}
	}
	\subfloat[VGG16]{
	\includegraphics[scale=0.35]{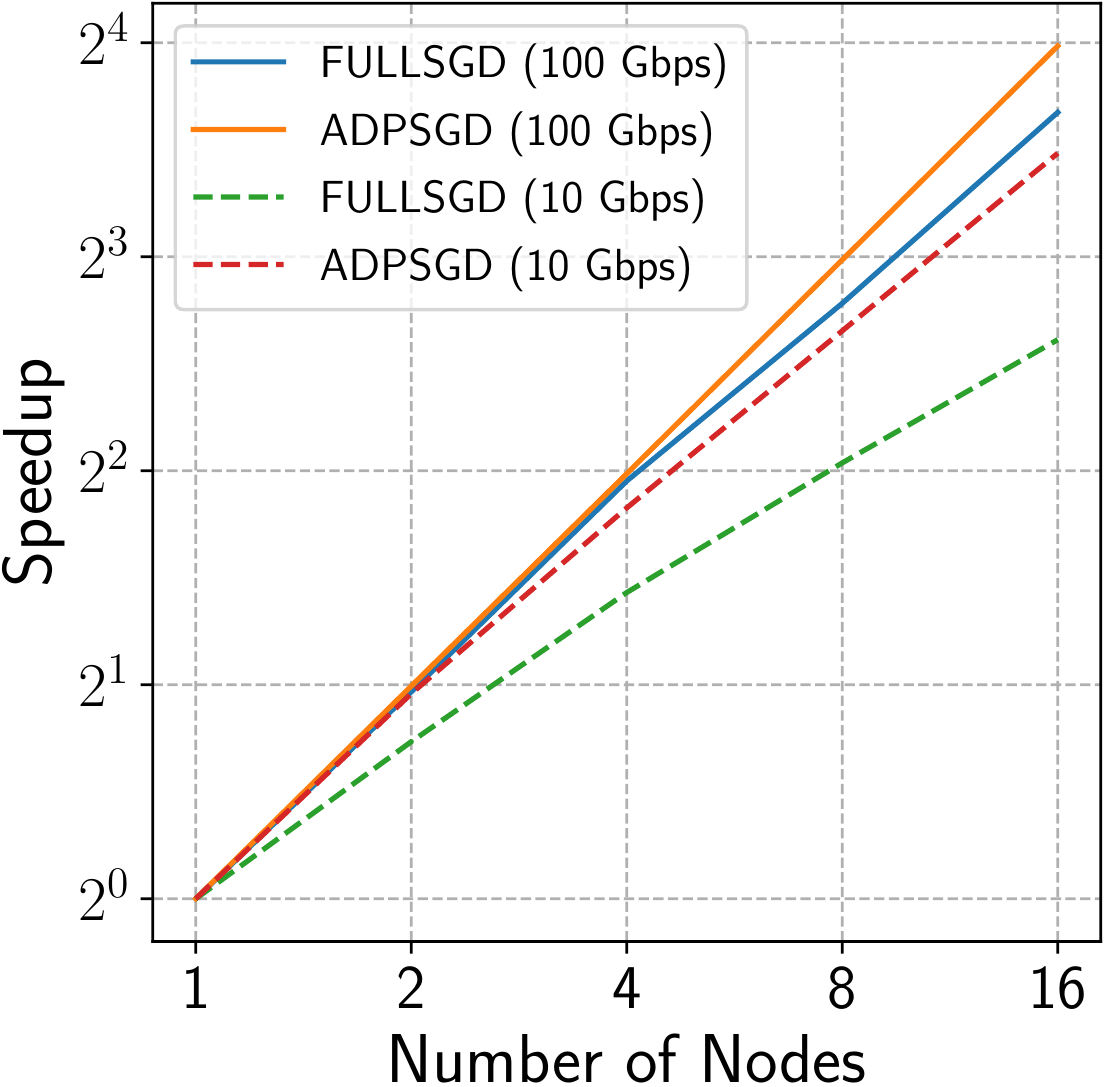}
		\label{fig:speedupsvgg}
	}
	\vspace{-0.5em}
	\caption{Speedups against single-node vanilla SGD}
		\vspace{-0.8em}
	\label{fig:speedups}
\end{figure}

Figure~\ref{fig:googlenettime} and~\ref{fig:vggtime} show the computation and communication time of different versions of SGD. 
We conduct the experiments with two bandwidth configurations. 
The first one is the original 100Gbps InfiniBand connection that reflects HPC clusters. 
The second one is an emulated 10Gbps connection, which is common in cloud settings  (we use {\em trickle} to throttle the download and upload rate of each node to 5Gbps).  
Among different versions, our ADPSGD has the smallest communication overhead, though it incurs a small amount of extra overhead in computation. 
The extra overhead is due to the computation of $S_k$ in Algorithm~\ref{alg2}; however, it cost less than $1\%$ of the original computation.  
Comparing the total execution time, our ADPSGD achieves $1.14$x and $1.24$x speedups against FULLSGD  for training GoogLeNet and VGG16 with 100Gbps connection. 
The speedups increase to $1.46$x and $1.83$x with 10Gbps connection.

To show how our ADPSGD improves the scalability of distributed training, we run FULLSGD and ADPSGD on $2$, $4$, $8$ and $16$ GPUs, and compare their total execution time with single-node vanilla SGD. 
Figure~\ref{fig:speedups} shows the speedups of distributed FULLSGD and ADPSGD against single-node vanilla SGD. 
For GoogLeNet, because most of the execution time are spent on computation, the speedups of FULLSGD with both 100Gbps and 10Gbps connections are acceptable; however, our ADPSGD is still beneficial and achieves almost linear speedups over 16 nodes. 
For VGG16, because communication is relatively expensive, it becomes a bottleneck in FULLSGD. 
As shown in Figure~\ref{fig:speedupsvgg}, FULLSGD on $16$ nodes achieves $12.77$x speedup against single-node SGD when the connection has 100Gbps bandwidth, and the speedup decreases to $6.12$x when the bandwidth is throttled to 10Gbps. 
Our ADPSGD effectively reduces the communication overhead and achieves almost linear speedups over 16 nodes.


\subsection{Results on ImageNet}

\begin{figure*}
	\centering
	\subfloat[Training Loss]{
	\includegraphics[scale=0.33]{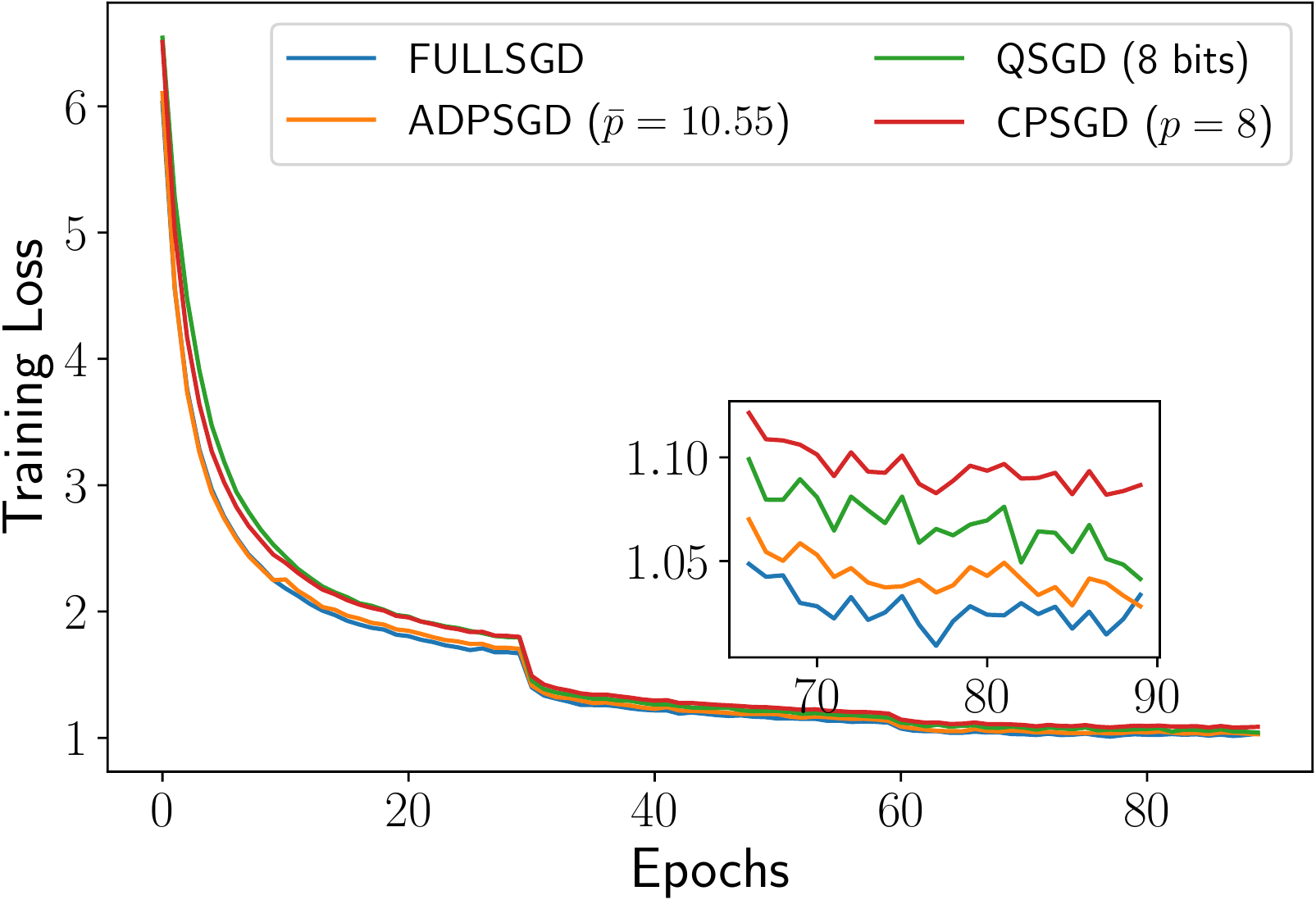}
	}
	\subfloat[1-Crop Validation Accuracy]{
	\includegraphics[scale=0.33]{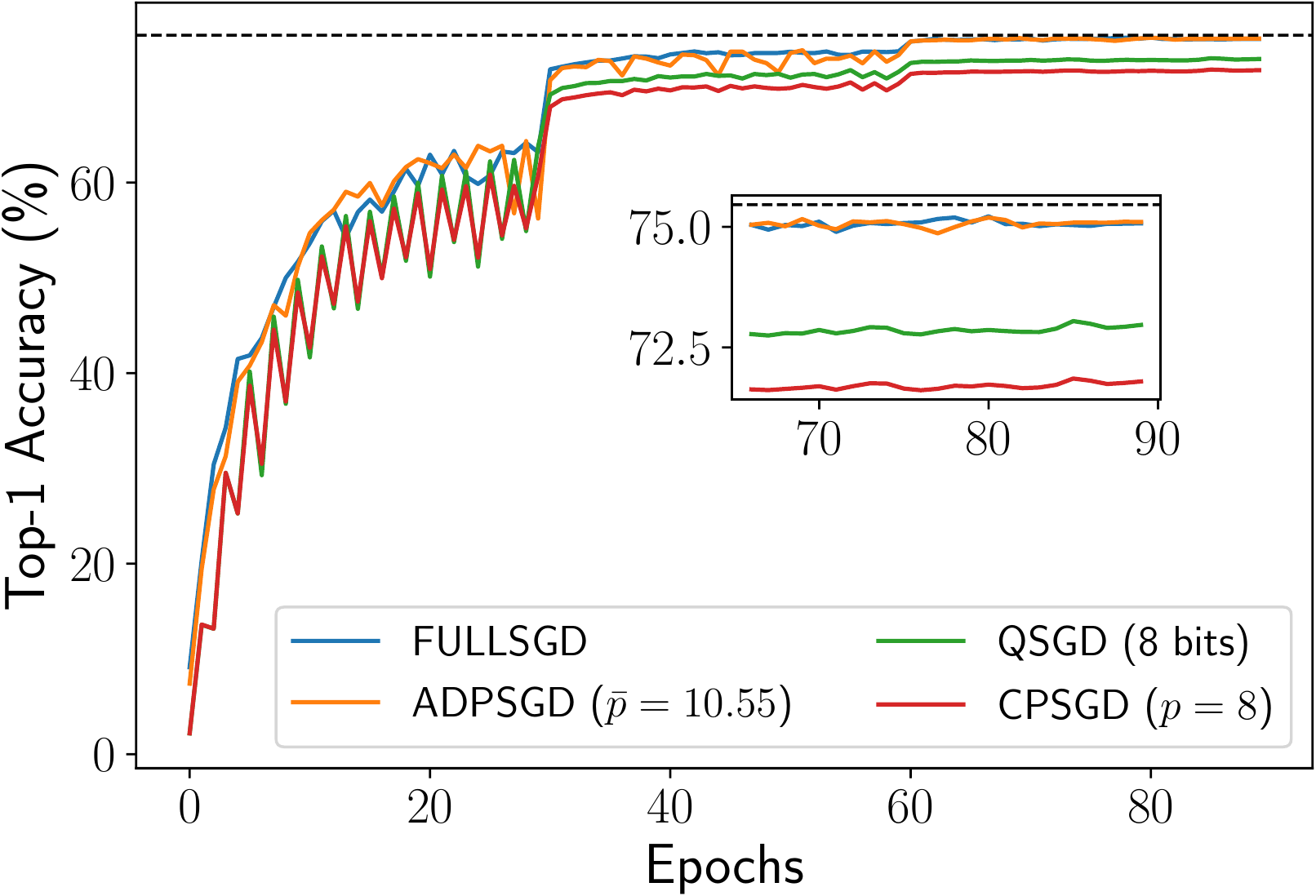}
			\label{fig:imagenetresnetacc}
	}
	\subfloat[Execution Time]{
	\includegraphics[scale=0.34]{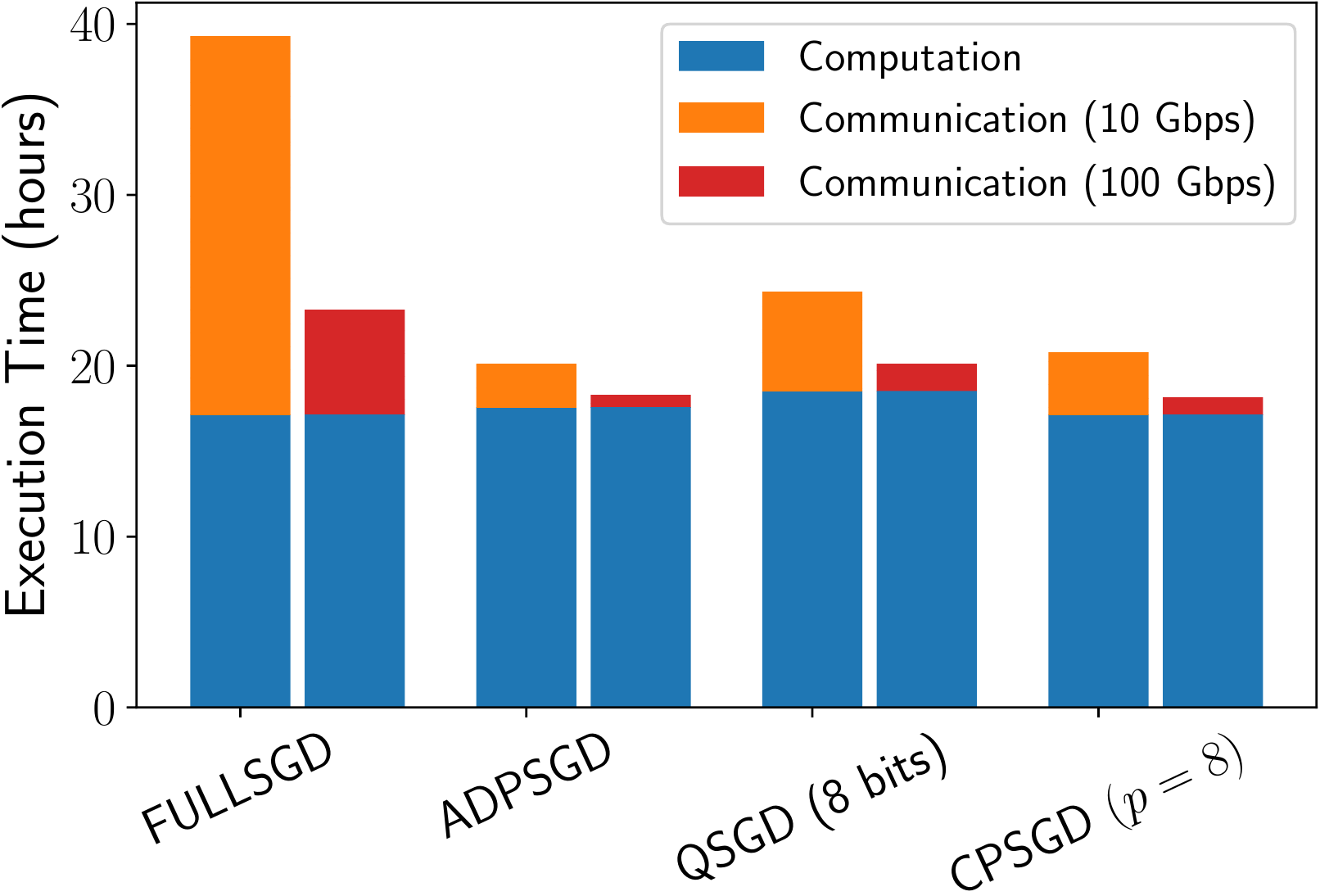}
			\label{fig:imagenetresnettime}
	}
	\vspace{-0.5em}
	\caption{Training ResNet50 on ImageNet with 16 GPUs}
		\vspace{-1em}
		\label{fig:imagenetresnet}
\end{figure*}

\begin{figure*}[t]
	\centering
	\subfloat[Training Loss]{
	\includegraphics[scale=0.33]{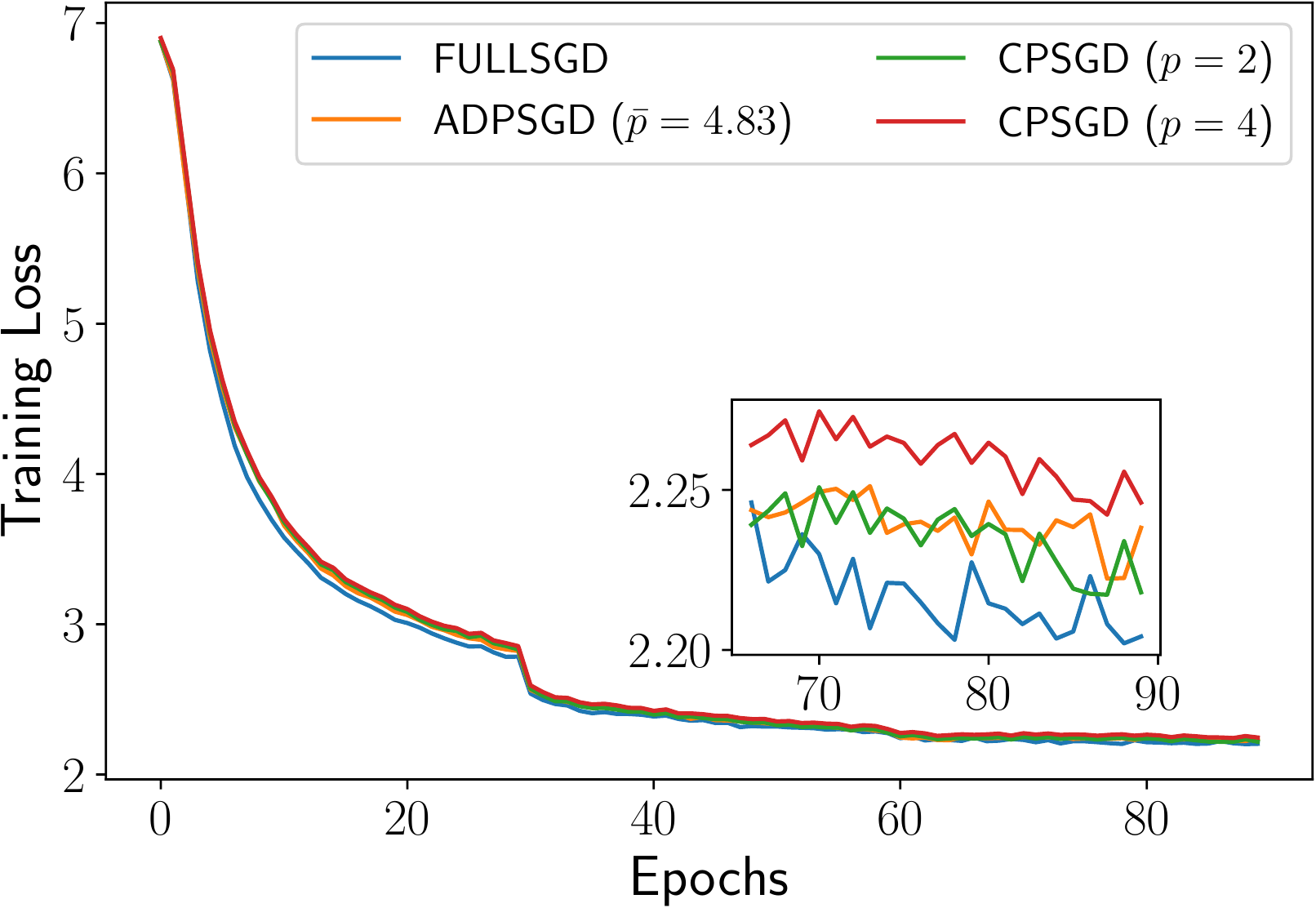}
	}
	\subfloat[1-Crop Validation Accuracy]{
	\includegraphics[scale=0.33]{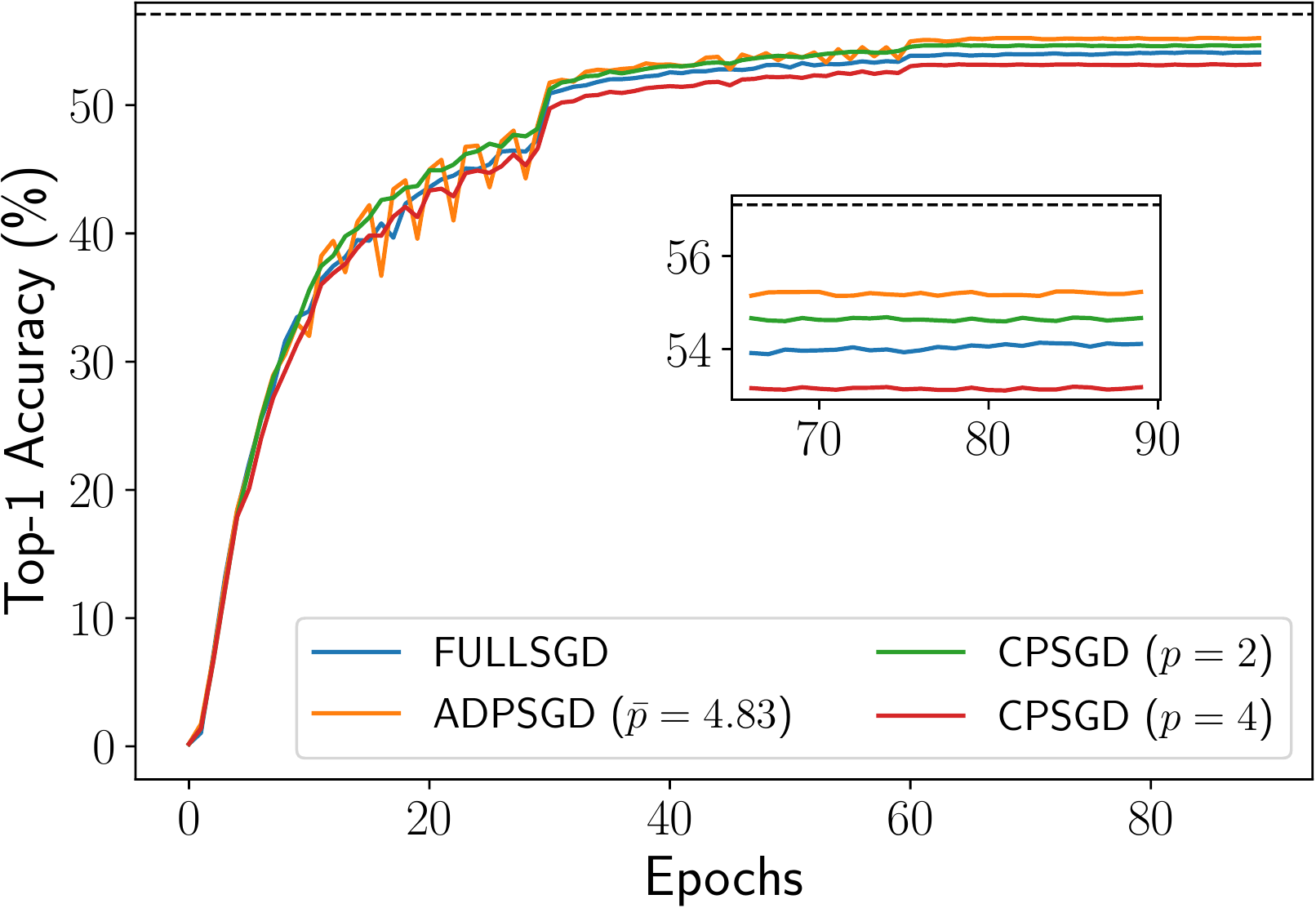}
			\label{fig:alexnet_imagenetresnetacc}
	}
	\subfloat[Execution Time]{
	\includegraphics[scale=0.34]{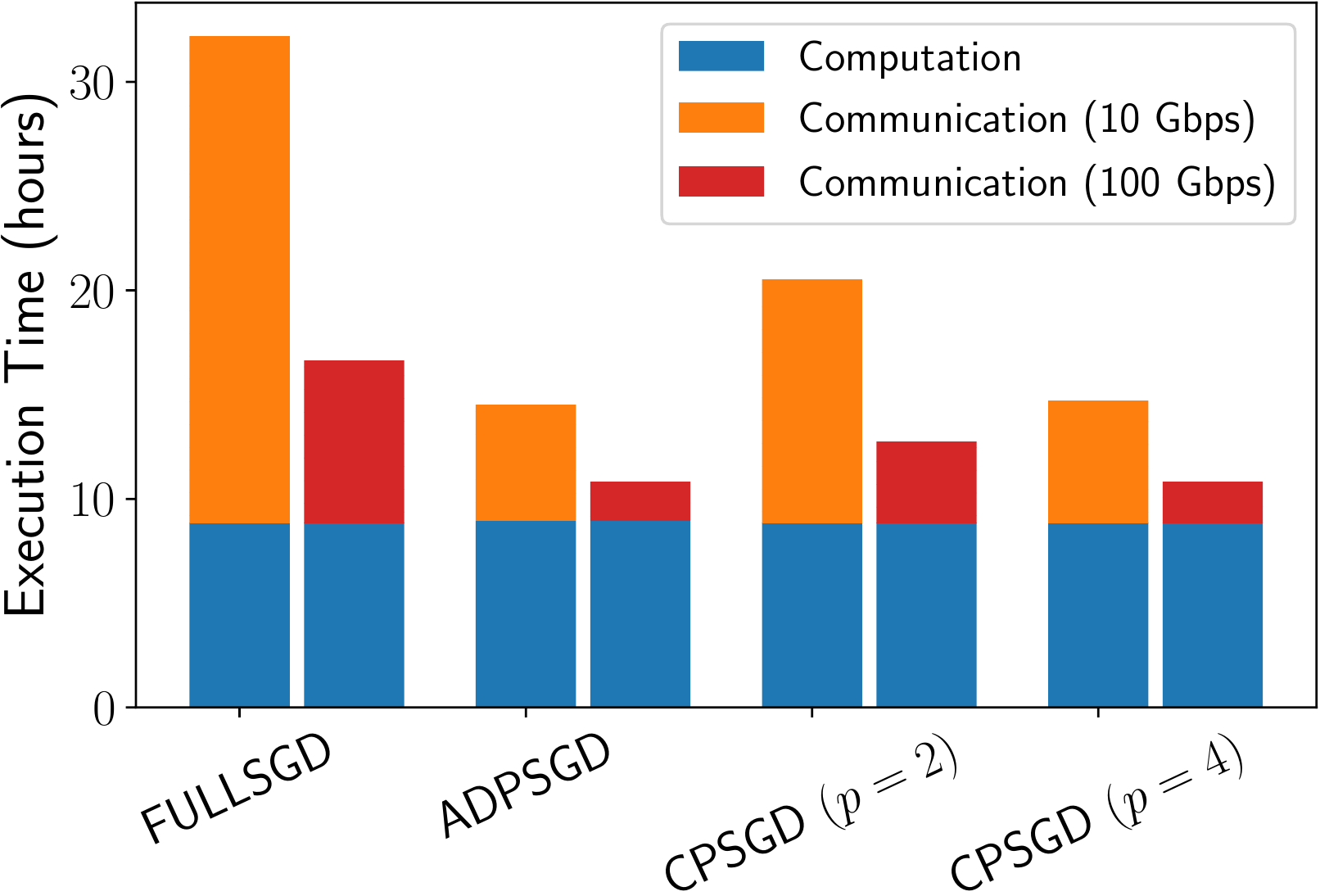}
			\label{fig:alexnet_imagenetresnettime}
	}
	\caption{\revision{Training AlexNet on ImageNet with 16 GPUs}}
		\label{fig:alexnet_imagenetresnet}
\end{figure*}

ILSVRC 2012 (ImageNet)~\cite{imagenet_cvpr09} is a much larger dataset than CIFAR-10. 
 There are a total of 1,281,167 images in 1000 classes for training, and 50,000 images for validation. 
 The images vary in dimensions and resolution. The average resolution is 469x387 pixels. 
Normally the images are resized to 256x256 pixels for neural networks.

We first train a ResNet50~\cite{DBLP:journals/corr/HeZRS15} on ILSVRC 2012. 
We apply the {\em linear scaling rule} and the {\em gradual warmup} techniques proposed in 
literature~\cite{DBLP:journals/corr/GoyalDGNWKTJH17} for all versions of SGD in our experiments. 
Specifically, the first $8$ epochs are warmup -- the learning rate is initialized to $0.1$ and is  
increased by $0.1$ in each epoch until the epoch $8$. 
Starting from the epoch $8$, the learning rate is fixed at $0.8$ until epoch $30$ when the learning rate is decreased to $0.08$. 
At epoch $60$, the learning rate is further decreased to $0.008$. 
The total number of epochs is $90$. 
For ADPSGD and CPSGD, the periodic averaging is applied after the warmup phase (i.e., the first $8$ epochs in ADPSGD and CPSGD are the same as in FULLSGD).  
For ADPSGD, Algorithm~\ref{alg2} is applied with $p_{init}=4$, $K_s=0.2K$ where $K$ is the total number of training iterations. 
For QSGD, the gradient quantization is also started after the warmup phase.

Figure~\ref{fig:imagenetresnet} shows the training loss and the top-1 validation accuracy achieved by different versions. (We evaluate the accuracy on the validation data in ILSVRC 2012).  
The validation images are resize to $256\times 256$ in which the center $224\times 224$ crop are used for validation. (This is the same as in the original ResNet paper~\cite{DBLP:journals/corr/HeZRS15}). 
We can see that our ADPSGD achieves smaller training loss and higher validation accuracy than CPSGD with $p=8$. 
The average averaging period in our ADPSGD is $10.55$, which means that our ADPSGD achieves better convergence than CPSGD while requiring  less communication. 
Compared with QSGD, our ADPSGD also achieves  smaller training loss and higher validation accuracy while requiring less communication. 
Compared with FULLSGD, our ADSGD achieves slightly larger training loss but almost the same test accuracy. 
The dashed line in Figure~\ref{fig:imagenetresnetacc} is the accuracy achieved by small-batch SGD -- we train ResNet50 with batch size $256$ and initial learning rate $0.1$, and the best accuracy we achieved is $75.458\%$. 
The validation error (i.e., $1-accuracy$) reported in the original ResNet paper~\cite{DBLP:journals/corr/HeZRS15} is $24.7\%$, which is close to the accuracy we achieved, so we use $75.458\%$ as the baseline accuracy for comparison. 
We can see in Figure~\ref{fig:imagenetresnetacc} that the best accuracy achieved by our ADPSGD is quite close to the baseline. 

Figure~\ref{fig:imagenetresnettime} shows the execution time of different versions. 
About $25\%$ of total execution time of FULLSGD is spent on communication with 100Gbps connection, and the ratio increases to $56\%$ with 10Gbps connection. 
Our ADPSGD effectively reduces the communication overhead and achieves $1.27x$ and $1.95$x speedups against  
FULLSGD with 100Gbps and 10Gbps connections,  respectively. 
The speedups of 16-node ADPSGD over single-node vanilla SGD are $15.75$x and $14.94$x with 100Gbps and 10Gbps connections respectively, indicating that our ADPSGD achieves linear speedup across 16 nodes.

We also train an AlexNet~\cite{NIPS2012_4824} on ILSVRC 2012. 
We use the same warmup and learning rate schedule as for ResNet50 above. 
The configurations for ADPSGD is also the same.  
Figure~\ref{fig:alexnet_imagenetresnet} shows the results of training AlexNet on ImageNet dataset. 
The dashed line in Figure~\ref{fig:alexnet_imagenetresnetacc} shows the baseline top-1 accuracy ($57.08\%$) achieved by single-node training (i.e., with batch-size $128$).  
The accuracy is close to those reported in~\cite{DBLP:journals/corr/Krizhevsky14} and~\cite{NIPS2017_6770}.  
There is a gap between the accuracy achieved by large-batch SGD and the baseline accuracy. Hoffer {\em et al.}~\cite{NIPS2017_6770} attribute this gap to insufficient exploration of the parameter space and show that the gap will be reduced by training more iterations. 
We do not increase the total number of iterations in our experiment for keeping fixed computation complexity. 
The results confirm the advantage of our ADPSGD over FULLSGD and CPSGD as it consistently achieves higher accuracy than other versions. 


%% file: text/discuss.tex
\section{Discussion}
\label{sec:twlbs}

\subsection{Improved Generalization for Large-Batch Training}
A well-known problem of training with large batches is the  
{\em generalization gap}~\cite{NIPS2017_6770, l.2018a, DBLP:journals/corr/KeskarMNST16, LeCun1998}. 
That is, when a large batch size is used while training deep neural networks, the trained models usually have lower test accuracy values than the models trained with small batches.  
 This prevents us from scaling out mini-batch SGD to a large number of  nodes.  
 While we focus on reducing the communication overhead in this work, we observed in 
 the previous section  that  ADPSGD achieves higher or equal test accuracy compared with  
the full-communication SGD. 
Zhou {\em et al.}~\cite{DBLP:journals/corr/abs-1708-01012} also observed this phenomenon. 
However, as suggested by the convergence analysis in \S\ref{sec:analysis}, periodic parameter averaging SGD  converges slower than full-communication SGD because of the variance of model parameters among the machines. 
This point is also validated by our experiments, as periodic parameter averaging SGD has larger training losses despite higher accuracy values.  

Therefore, we argue that the improved generalization achieved by periodic parameter averaging SGD actually comes from its higher potential to escape sharp minima and avoid overfitting during the training process. 
Because explaining the generalization of neural networks is still an open problem, we provide an intuitive explanation of ADPSGD over large-batch SGD based on this argument.

A popular hypothesis for the decreased accuracy  of large-batch SGD is that 
the  gradients computed on 
large mini-batch are close to the actual gradients and the steps make the algorithm quickly converge to and  
be trapped in a local sharp minimum~\cite{DBLP:journals/corr/KeskarMNST16, pmlr-v70-dinh17b, NIPS2017_6770, l.2018dont}. 
In contrast, the large gradient estimation noise in small mini-batches encourages the model parameters to  
escape  sharp minima, thereby leading to more efficient {\em sampling of the parameter space}~\cite{DBLP:journals/corr/KeskarMNST16}. 
Based on this hypothesis, techniques for escaping the sharp minima by injecting random noise to the training  
process have been proposed~\cite{Welling:2011:BLV:3104482.3104568, DBLP:journals/corr/abs-1805-07898, DBLP:journals/corr/ChaudhariCSL16, JMLR:v15:srivastava14a}. 
 In this sense, our ADPSGD adds noise to the training process by allowing a small deviation of the  
trajectories of model parameters on different nodes. 
 Intuitively, because the model parameters on each node are updated with gradients computed on smaller  
mini-batches, they have higher potential to escape sharp minima. 
And,  because the parameter averaging is performed once in a few iterations, it is likely that the model  
parameters on certain nodes have escaped the sharp minima and they can drag the model parameters on  
other nodes out of the sharp minima. 

It is worth noting that, although periodic parameter averaging is reminiscent of using  larger batches, our algorithm adds noise to the training process without changing the learning rate while one needs to increase the learning rate to keep the noise scale when using a larger batch size~\cite{DBLP:journals/corr/abs-1711-00489}.   
As we described in the background (\S\ref{sec:back2}), there is a theoretical limit of the learning rate that depends on the smoothness of the objective function. 
Therefore, linear scaling of the learning rate with the batch size is not a universally effective rule to overcome generalization gap of large-batch training~\cite{DBLP:journals/corr/abs-1805-07898, NIPS2017_6770}.

Although periodic averaging has higher potential to escape sharp minimum than full-communication SGD,  
the analysis in \S\ref{sec:analysis} indicates that a larger averaging period slows down the convergence of the algorithm.    
The conflict between the sampling efficiency and the convergence rate suggests that there is an optimal averaging period for SGD to achieve the best accuracy.  
This explains the high accuracy of our ADPSGD as it has the advantage of escaping sharp minimum  while preserving a fast convergence rate.

\subsection{Pitfalls of Oversimplified Analysis}
\label{sec:pitfalls}
 Wang {\em et al.}~\cite{DBLP:journals/corr/abs-1810-08313} 
 have worked on the same problem but proposed an approach that can be 
 viewed as the opposite to ours. 
They argue that periodic parameter averaging SGD should use a large averaging periodic at first and gradually reduce the averaging period during the training process. 
We now point out the errors in their argument, and thereby justify the method presented  in this paper. 

In their analysis (see (56) in~\cite{DBLP:journals/corr/abs-1810-08313}), they derive:
\begin{flalign}
\label{eq:wangeq1}
	\sum_{k=pt}^{p(t+1)-1}&\mathbb{E}\norm{\nabla f\myparam{W_k\dotavg{n}}} \leq \nonumber \\ &\frac{2\myparam{\mathbb{E}f\myparam{W_{pt}\dotavg{n}} - \mathbb{E}f\myparam{W_{p(t+1)}\dotavg{n}}}}{\gamma_{pt}} \nonumber \\ &- \myparam{\frac{1-O(p^2)}{n}}\sum_{k=pt}^{p(t+1)-1}\mathbb{E}\normF{G(W_k)} + O(p^2) 
\end{flalign} 
where $G(W_k)= \myrect{\nabla f_1(w_{k,1}),...,\nabla f_n(w_{k,n}) }$ are the  gradients on $n$ nodes  computed with local data. 
This can been seen by moving the first term on the right hand side of (56) in~\cite{DBLP:journals/corr/abs-1810-08313} to the left hand side (they denote the communication period $p$ as $\tau_j$). 
This step is correct and can also be derived from (\ref{eq:step8}) in our paper.  
However, the authors oversimplify this bound by assuming $O(p^2)\leq 1$ and removing the second term on the right hand of (\ref{eq:wangeq1}). (See (57) and (58) in~\cite{DBLP:journals/corr/abs-1810-08313}) 
The authors conclude that $p$ can be decreasing as the first term decreases over iterations, but they completely ignore the fact that $\mathbb{E}\normF{G(W_k)}$ is decreasing. 
If using a small $p$ in early iterations when $\mathbb{E}\normF{G(W_k)}$ is large, the average gradient norm will have a much smaller bound, which means a faster convergence of the algorithm. 
This confirms our conclusion that using small communication period in early phase of the training process is more beneficial than in later phase. 

To validate that decreasing communication period is not helpful, we test periodic parameter averaging SGD that communicates every 20 iterations in the first 80 epochs and every 5 iterations in the remaining 80 epochs for training GoogleNet and VGG16 on CIFAR10.  
Because it incurs $500$ ($2000\div 20 + 2000 \div 5$) synchronizations, its communication overhead is the same as CPSGD with $p=4000\div 500=8$.
For GoogleNet, the smallest training loss it achieves is $0.023$, which is one order of magnitude larger than the smallest training losses of other versions in Figure~\ref{fig:googlenetloss}. 
For VGG16, the smallest training loss it achieves is $0.15$, which is also much larger than the smallest losses of other versions in Figure~\ref{fig:vggloss}.  
The best test accuracy it achieves for the two models are $91.84$ and $89.78$, which are lower than the best accuracies of other versions in Figure~\ref{fig:googlenetacc} and~\ref{fig:vggacc}.  


%% file: text/related.tex
\section{Related Work}
Many techniques have been proposed to reduce the communication overhead in distributed training.

\textbf{Gradient Compression. } 
A popular approach to reducing communication  overhead of distributed training is to perform {\em compression}  of the gradients~\cite{lin2018deep}\cite{DBLP:journals/corr/AjiH17}\cite{NIPS2017_6749}\cite{NIPS2017_6768}\cite{1-bit}. 
For example, Wen {\em et al.}~\cite{NIPS2017_6749} propose to  {\em quantize} the gradients to three numerical levels, and thus only two bits are transmitted for a gradient value instead of 32 bits.  
Lin {\em et al.}~\cite{lin2018deep} claim that by combining multiple compression techniques, they can achieve 270x to 600x compression ratio without losing accuracy. 
 Strom~\cite{Strom2015ScalableDD} proposed to only send gradient components larger than a predefined threshold.  
Aji {\em et al.}~\cite{DBLP:journals/corr/AjiH17} presented a heuristic approach to truncate the smallest gradient components and only communicate the remaining large ones. 
Alistarh {\em et al.}~\cite{NIPS2017_6768} proposed a quantization method named QSGD and gave its convergence rate for both convex and non-convex optimization.  
\revision{Finally, from the system side, Renggli {\em et al.}~\cite{renggli2019sparcml}  have developed new  communication primitives 
for dealing with  compressed and/or sparse data.} 

Despite their popularity in research, these compression-based approaches yield small practical gains on HPC clusters with fast connections due to several reasons. 
First, gradient compression commonly assume that the model parameters are maintained by one or multiple {\em parameter servers}~\cite{NIPS2014_5597}, and they cannot be easily combined with bandwidth-optimal Allreduce, which is a more efficient way to aggregate data from multiple machines~\cite{Patarasuk:2009:BOA:1482176.1482266}.  
Second, these quantization-based methods usually change the convergence property of SGD as some information of the gradients is lost in compression~\cite{NIPS2018_7519}. 
Third, the  compression or quantization procedure  itself incurs computation overheads, which can defeat the benefits of saved communication time, especially when the interconnect is fast. 

\textbf{Periodic Averaging SGD.} 
Periodic averaging has been widely used to reduce the communication overhead in distributed training~\cite{JMLR:v15:hazan14a}\cite{Johnson:2013:ASG:2999611.2999647}\cite{DBLP:journals/corr/SmithFMTJJ16}\cite{2016arXiv160607365Z}\cite{45187}\cite{2017arXiv170206269W}. 
The idea is to simply perform the synchronization only once in a few iteration. 
Earlier works demonstrate that periodic averaging can be incorporated into asynchronous SGD~\cite{Dean:2012:LSD:2999134.2999271}\cite{Zhang:2015:DLE:2969239.2969316}. 
Downpour SGD~\cite{Dean:2012:LSD:2999134.2999271} maintains the model parameters on each node locally; 
after $k$ iterations of local updates, the change of the local model parameters  is sent to the parameter server asynchronously, and the parameter serve updates the global model parameter according to the received parameter changes. 
Zhang {\em et al.}~\cite{Zhang:2015:DLE:2969239.2969316} improve Downpour SGD by presenting EASGD which simulates an elastic force to link the model parameters on different nodes with the global model parameters. 
They show that EASGD with periodic communication has guaranteed convergence on quadratic and strongly-convex optimization if the hyperparamters are properly configured. 
Recently, Zhou {\em et al.}~\cite{DBLP:journals/corr/abs-1708-01012} prove that $O(1/\sqrt{MK})$ convergence rate can be achieved by constant periodic averaging SGD for non-convex optimization. 
However, they  claim that it  
is hard to determine the optimal period in practice because it depends on  
certain unknown quantities  associated with  the objective function. 
It is  also unclear from their analysis how one can adjust the averaging period to achieve the  
optimal convergence. 
\revision{ Haddadpour {\em et al.}~\cite{haddadpour2019local} studied the convergence of local SGD and gave an upper bound of communication period for non-convex optimization under the PL condition~\cite{karimi2016linear}.  
Their theoretical analysis also suggests that the  communication period should be increasing.  
However, because of the stronger assumption, their work is only evaluated with a logistic regression model and is not applied to neural networks. }

%% file: text/conclusion.tex
\section{Conclusion}
In this paper, we have presented an adaptive period scheduling method for periodic parameter averaging SGD. 
We demonstrate that our adaptive periodic parameter averaging SGD achieves better convergence than constant periodic averaging SGD while requiring the same or even smaller amount of communication. 
Our experiments with image classification benchmarks show that our approach indeed achieves smaller training loss and higher test accuracy than different configurations of constant periodic averaging. 
Compared with full-communication SGD, our method achieves 1.14x to 1.27x speedups with a 100Gbps connection, and 1.46 to 1.95x speedups with an emulated 10Gbps connection, leading to linear speedups against single-node SGD across 16 GPUs.  
Compared with gradient-quantization SGD using 8 bits, our algorithm achieves faster convergence with only half of the communication.